\documentclass{article}

\usepackage{microtype}
\usepackage{graphicx}
\usepackage{subcaption}
\usepackage{booktabs} 

\usepackage{hyperref}
\usepackage{url}
\usepackage{xcolor}
\usepackage{multirow}
\usepackage{adjustbox}
\usepackage{color, colortbl}
\definecolor{Gray}{gray}{0.9}

\usepackage[preprint]{icml2026}

\usepackage{amsmath}
\usepackage{amssymb}
\usepackage{mathtools}
\usepackage{amsthm}
\usepackage{amsfonts}
\usepackage{bm}
\usepackage[most]{tcolorbox}
\usepackage{makecell}

\usepackage{wrapfig}

\usepackage[font=small,skip=0pt]{caption}
\usepackage{listings}
\definecolor{codegreen}{rgb}{0,0.6,0}
\definecolor{codegray}{rgb}{0.5,0.5,0.5}
\definecolor{codepurple}{rgb}{0.58,0,0.82}
\definecolor{backcolour}{rgb}{0.95,0.95,0.92}
\lstdefinestyle{mystyle}{
    backgroundcolor=\color{backcolour},   
    commentstyle=\color{codegreen},
    keywordstyle=\color{magenta},
    numberstyle=\tiny\color{codegray},
    stringstyle=\color{codepurple},
    basicstyle=\ttfamily\footnotesize,
    breakatwhitespace=false,         
    breaklines=true,                 
    captionpos=b,                    
    keepspaces=true,                 
    numbers=left,                    
    numbersep=5pt,                  
    showspaces=false,                
    showstringspaces=false,
    showtabs=false,                  
    tabsize=2
}
\definecolor{myhighlight}{RGB}{220,240,255}

\usepackage{pifont}
\newcommand{\myheight}{\vphantom{d_{\mathrm{TV}}(\mathbb{P}_i, \mathbb{P}_k)}}

\definecolor{babypink}{rgb}{0.96, 0.76, 0.76}
\definecolor{champagne}{rgb}{0.97, 0.91, 0.81}

\usepackage{listings}
\definecolor{codegreen}{rgb}{0,0.6,0}
\definecolor{codegray}{rgb}{0.5,0.5,0.5}
\definecolor{codepurple}{rgb}{0.58,0,0.82}
\definecolor{backcolour}{rgb}{0.95,0.95,0.92}
\lstdefinestyle{mystyle}{
    backgroundcolor=\color{backcolour},   
    commentstyle=\color{codegreen},
    keywordstyle=\color{magenta},
    numberstyle=\tiny\color{codegray},
    stringstyle=\color{codepurple},
    basicstyle=\ttfamily\footnotesize,
    breakatwhitespace=false,         
    breaklines=true,                 
    captionpos=b,                    
    keepspaces=true,                 
    numbers=left,                    
    numbersep=5pt,                  
    showspaces=false,                
    showstringspaces=false,
    showtabs=false,                  
    tabsize=2
}

\usepackage{hyperref}

\usepackage[capitalize,noabbrev]{cleveref}

\theoremstyle{plain}
\newtheorem{theorem}{Theorem}[section]

\theoremstyle{definition}

\newtheorem{assumption}[theorem]{Assumption}
\theoremstyle{remark}
\newtheorem{remark}[theorem]{Remark}

\usepackage[textsize=tiny]{todonotes}

\icmltitlerunning{VAO: Validation-Aligned Optimization for Cross-Task Generative Auto-Bidding}

\begin{document}

\twocolumn[
  \icmltitle{VAO: Validation-Aligned Optimization for Cross-Task Generative Auto-Bidding}



  \icmlsetsymbol{equal}{*}

  \begin{icmlauthorlist}
    \icmlauthor{Yiqin Lv}{equal,comp,sch,yyy}
    \icmlauthor{Zhiyu Mou}{equal,comp}
    \icmlauthor{Miao Xu}{comp}
    \icmlauthor{Jinghao Chen}{sch}
    \icmlauthor{Qi Wang}{sch}
    \icmlauthor{Yixiu Mao}{sch}
    \icmlauthor{Yun Qu}{sch}
    \icmlauthor{Rongquan Bai}{comp}
    \icmlauthor{Chuan Yu}{comp}
    \icmlauthor{Jian Xu}{comp}
    \icmlauthor{Bo Zheng}{comp}
    \icmlauthor{Xiangyang Ji}{sch}
  \end{icmlauthorlist}

  \icmlaffiliation{yyy}{Work was done during a visit to Tsinghua University and an internship at Alibaba Group}
  \icmlaffiliation{comp}{Alibaba Group, Beijing, China}
  \icmlaffiliation{sch}{Department of Automation, Tsinghua University, Beijing, China}

  \icmlcorrespondingauthor{Yiqin Lv}{lvyiqin25@gmail.com}

  \icmlkeywords{Machine Learning, ICML}

  \vskip 0.3in
]



\printAffiliationsAndNotice{\icmlEqualContribution}

\begin{abstract}
Generative auto-bidding has demonstrated strong performance in online advertising, yet it often suffers from data scarcity in small-scale settings with limited advertiser participation.
While cross-task data sharing is a natural remedy to mitigate this issue, naive approaches often introduce gradient bias due to distribution shifts across different tasks, and existing methods are not readily applicable to generative auto-bidding.
In this paper, we propose Validation-Aligned Optimization (VAO), a principled data-sharing method that adaptively reweights cross-task data contributions based on validation performance feedback.
Notably, VAO aligns training dynamics to prioritize updates that improve generalization on the target task, effectively leveraging auxiliary data and mitigating gradient bias.
Building on VAO, we introduce a unified generative autobidding framework that generalizes across multiple tasks using a single model and all available task data.
Extensive experiments on standard auto-bidding benchmarks validate the effectiveness of our approach.

\end{abstract}

\section{Introduction}
\begin{figure}[t]
    \centering
    \includegraphics[width=1.0\linewidth]{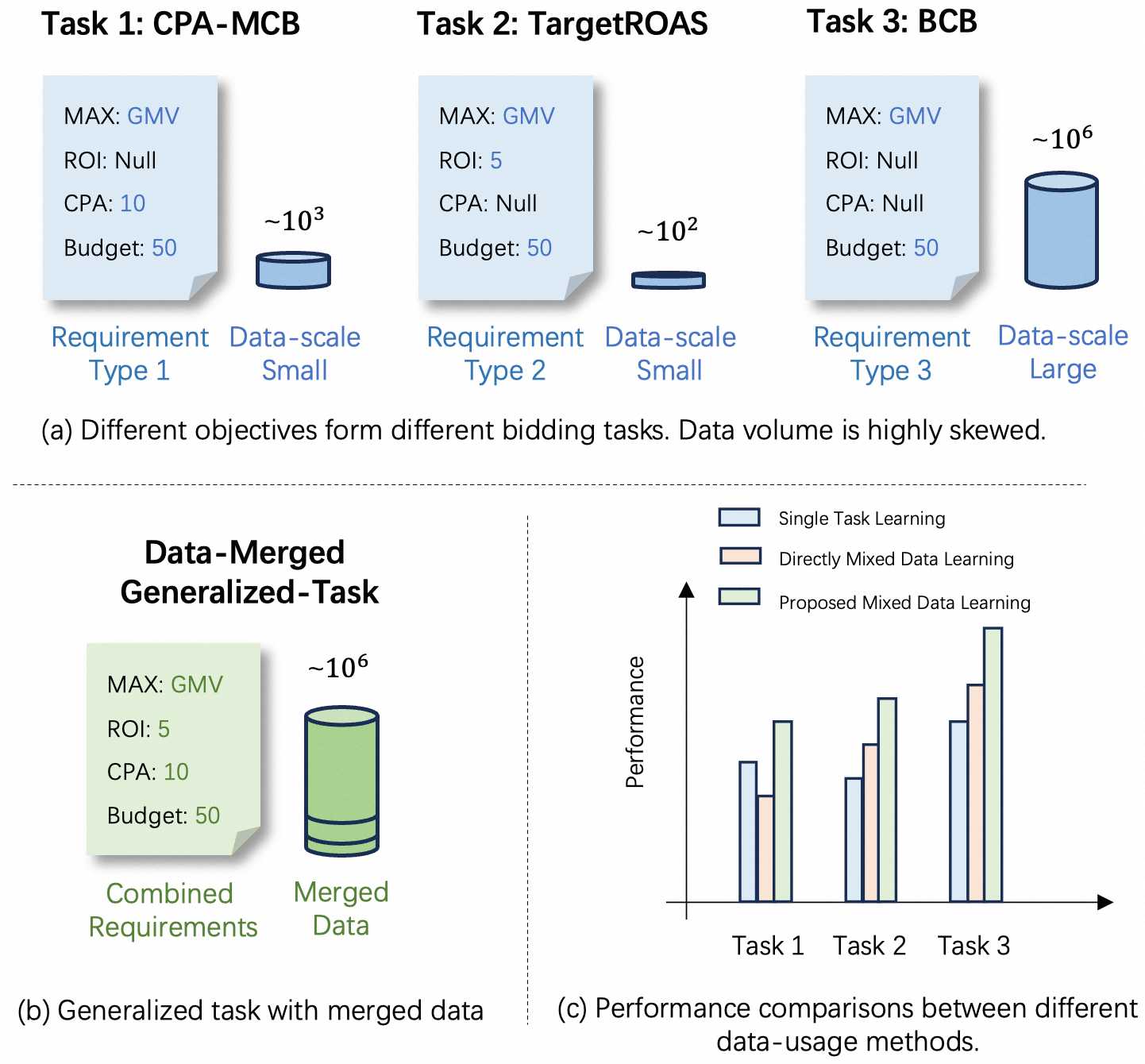}
    \caption{ 
    Consider three representative auto-bidding tasks characterized by varying data densities: BCB, a high-quantity task focused on maximizing Gross Merchandise Value (GMV) within budget constraints, and two data-scarce tasks: TargetROAS and CPA-MCB. The latter ones introduce auxiliary return-on-investment (ROI) and cost-per-action (CPA) constraints, respectively.
    }
    \label{fig:data_imbalanced}
\end{figure}

In modern online advertising, auto-bidding has become a cornerstone of campaign optimization, effectively mapping high-level advertiser objectives to real-time bidding strategies \citep{He2021uscb}. 
To accommodate diverse advertising goals, platforms typically deploy specialized auto-bidding products for each objective, treating them as distinct tasks.
Prevailing paradigms rely on generative frameworks that train task-specific models using only corresponding historical data \cite{Guo2024aigb}.
However, real-world systems often face highly skewed data distributions across tasks. 
As illustrated in Fig.~\ref{fig:data_imbalanced}, 
While a few primary tasks possess abundant logs, the vast majority lie in the long tail, suffering from severe data scarcity due to limited advertiser participation. 
This sparsity poses significant challenges for bidding optimization, especially for generative approaches, which are highly sample-inefficient under sparse supervision.


Cross-task data sharing offers a natural and promising solution to this challenge \citep{yu2021conservative}. 
Despite the diversity of advertiser objectives, a key observation is the significant structural commonality in how objectives map to bidding trajectories. 
Specifically, since data across all tasks originates from the same platform and is governed by identical auction dynamics, the underlying mapping from bidding trajectories to their performance remains invariant.
This structural consistency suggests that knowledge from high-resource tasks can be effectively transferred to low-resource ones, providing a feasibility for cross-task data sharing.

\begin{figure}
    \centering
    \includegraphics[width=1.0\linewidth]{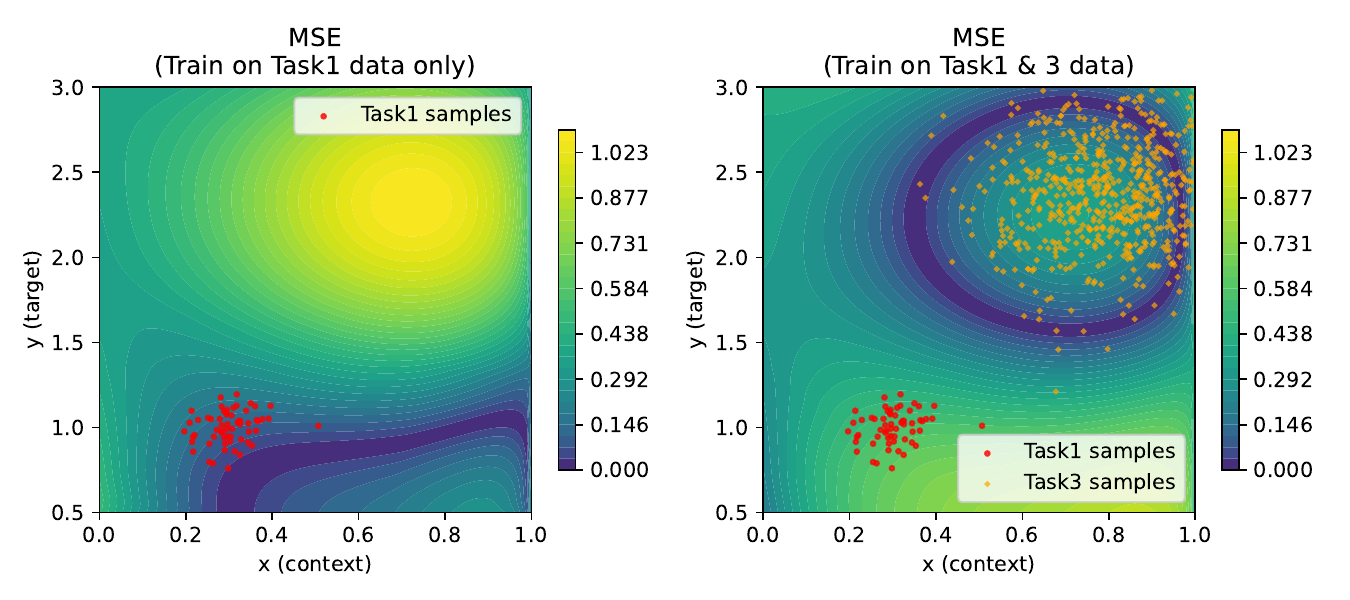}
    \caption{Direct cross-task data-sharing leads to performance degradation in data-scarce tasks.
    Specifically, Task 3 is high-resource with extensive data, whereas Task 1 is low-resource. }
    \label{fig:data_share_intro}
\end{figure}

However, direct data sharing across tasks can be detrimental due to the inherent distribution shift between objectives. 
Specifically, as illustrated in Fig. \ref{fig:data_share_intro}, we aim to optimize the low-resource Task 1 by directly leveraging the abundant data in Task 3. 
However, tasks with significantly larger data quantities often dominate the optimization landscape, leading to negative transfer or biased representations. This prioritization of majority patterns often leads to performance degradation for the minority tasks we intended to improve.
This phenomenon causes the model to overfit to dominant tasks and fail to generalize to tasks with sparse supervision.

Existing methods for mitigating negative transfer in cross-task data sharing are largely confined to reinforcement learning (RL) for auto-bidding frameworks. 
These methods typically rely on value-based signals or task-specific reward structures to guide transfer, addressing overestimation and exploitation biases in offline RL settings \citep{yu2021conservative,bai2024pessimistic}. 
However, in generative auto-bidding scenarios, the learning objective shifts toward matching complex trajectory distributions instead of simply optimizing a scalar value function. 
This misalignment presents an underexplored, technically challenging problem in data sharing.
In other words, it is difficult to achieve significant performance improvements on data-scarce tasks through effective data-sharing methods.




To this end, this work proposes Validation-Aligned Optimization (VAO), a cross-task data-sharing method for generative auto-bidding that enables robust performance uplift in data-scarce tasks. 
Under cross-task data sharing, we theoretically demonstrate that the generalization error of generative models is jointly determined by the training sample size and the discrepancy between the source and target distributions. 
This implies that naively increasing the available sample size from data sharing exacerbates distribution shifts and degrades generalization.  
To overcome the existing bottleneck, VAO proposes to estimate adaptive weights for source tasks from validation datasets to align with the optimization of target tasks. 
By leveraging augmented data while mitigating the negative effects of distribution shift, VAO can theoretically improve the generalization bound.
We further extend VAO to a practical multi-task setting by designing a unified generative auto-bidding architecture shared across all tasks.
Extensive experiments on standard auto-bidding benchmarks show that VAO consistently outperforms naive data sharing and state-of-the-art baselines, demonstrating its ability to leverage cross-task knowledge while maintaining target-specific fidelity.

\section{Related Works}
\label{sec:related_works}
The focus of this work is on cross-task data sharing in auto-bidding, which connects to but differs from existing fields.

\textbf{Multi-task learning (MTL)} trains shared models with adaptive weights to balance learning across tasks \citep{caruana1997multitask,kendall2018multi,chen2018gradnorm}.       
However, MTL optimizes for multi-task trade-offs, i.e., joint performance across all tasks, whereas our work optimizes for a specific target task by learning which auxiliary sources are beneficial for its generalization.
\textbf{Transfer learning} typically pre-trains on a source domain, then fine-tunes on a target \citep{weiss2016survey,jiang2022transferability,gholizade2025review}.
Our method differs by simultaneously mixing multiple heterogeneous sources with validation-aligned weights, which is essential when no single dominant source exists.
\textbf{Domain adaptation} reduces distribution shift between source and target by aligning feature distributions or learning domain-invariant representations \citep{jiang2022transferability,zhao2019learning,huang2025multi}.
Instead, we reweight source-task contributions based on validation performance without modifying the model's representations, providing a complementary approach that directly optimizes for target generalization.
In summary, the key distinction of VAO lies in its focus on \textit{adaptive cross-task data sharing}: rather than balancing multiple objectives or transferring from a single source, we learn task-specific mixture weights that maximize target performance in auto-bidding scenarios.
More discussions are attached in Appendix \ref{literature_review}.

\section{Preliminaries} \label{prelim_sec}
\subsection{Problem Setup}
Auto-bidding seeks a bidding policy that maximizes the cumulative value of impressions won over a finite bidding episode, e.g., one day \citep{He2021uscb,Mou2022sorl}.
Formally, the auto-bidding problem is usually modeled as a Markov Decision Process (MDP) defined by the tuple $<\mathcal{S}, \mathcal{A}, \mathcal{R},\mathcal{P}>$.
At each discrete time step $t\in[T]$, the state $\bm s_t\in\mathcal{S}$ describes the real-time advertising status, including the remaining time, left budget, consumption speed, etc.
The action $\bm a_t\in \mathcal{A}$ specifies a scaling factor applied to the bid at time $t$.
After taking action $\bm a_t$, the auto-bidding agent obtains a reward $r_t(\bm s_t,\bm a_t)\in\mathcal{R}$ reflecting the value of won impressions during $[t,t+1)$, and incurs a cost $c_t(\bm s_t,\bm a_t)$ that corresponds to the expenditure within this period.
The environment dynamics are characterized by $\mathcal{P}(\cdot|\bm s_t,\bm a_t)$ that governs the evolution of the state.

In online advertising, heterogeneous advertiser requirements lead to multiple auto-bidding tasks.
These tasks share the same objective of maximizing total cumulative rewards but differ in their constraints. 
Formally, each task $k\in\{1,\dots,K\}$ is characterized by a task-specific constraint function $\mathcal{G}_k(\tau)$, which defines the feasible region for a bidding trajectory $\tau_k=(\bm s_1, \bm a_1, \dots, \bm s_T)$. For instance, in the Budget-Constrained Bidding (BCB) setting \cite{wu2018budget},  the constraint function $\mathcal{G}_k(\tau)$ is instantiated as $\sum_{t=1}^Tc_t(\bm s_t,\bm a_t)-B\le0$, where $B$ denotes a pre-defined budget.
Formally, each task $k$ aims to seek an optimal trajectory that maximizes the expected cumulative reward under its task-specific constraint: $\max_{\tau_k} \;\mathbb{E}_{\tau_k\sim \mathcal{P}}\Big[\sum_{t=1}^T r_t(\bm s_t,\bm a_t)\Big], \quad \mathrm{s.t.} \quad \mathcal{G}_k(\tau)\le0$.


\subsection{Generative Auto-bidding}
Recent studies have demonstrated that the generative auto-bidding paradigm achieves state-of-the-art performance.
This is achieved by casting the auto-bidding problem as a conditional generative modeling task, thereby directly approximating the conditional distribution from offline datasets \citep{Guo2024aigb, li2025gas}.
Specifically, for each task $k$, the offline dataset $\mathcal{D}_k=\{(\tau^j_k,\bm y_k(\tau_k^j))\}_{j=1}^{N_k}$ consists of trajectories sampled from a task distribution $\mathbb{P}_k$.
Here, $\bm y_k$ is a conditioning vector that encapsulates desired performance targets and task-specific constraints (e.g., pre-defined budget or target ROI).
The objective of generative auto-bidding methods is:
\begin{equation}
\begin{aligned}
    \mathcal{L}_k(\mathcal{D}_k;{\bm\theta})=-\mathbb{E}_{(\tau_k,\bm y_k(\tau_k))\sim \mathcal{D}_k}[\log p_{\bm\theta}(\tau_k|\bm y_k(\tau_k)],
\end{aligned}   
\end{equation}
where $\bm\theta$ denotes the model parameters (with the task index $k$ omitted for brevity).


However, online advertising faces pervasive data scarcity across tasks due to limited advertiser participation.
Moreover, trajectories are typically collected under a single behavior policy, resulting in narrow and highly concentrated coverage in trajectory space.
This sparsity severely challenges generative auto-bidding methods, which are highly sample-inefficient under sparse supervision and thus struggle to generalize effectively \citep{belkhale2023data}.
From a principled perspective, let $\ell_k(\tau; \bm\theta)=-\log p_{\bm \theta}(\tau|\bm y_k(\tau))$ denote the per-trajectory loss for task $k$.
While the empirical gradient $\hat{\bm g}_k =\frac{1}{N_k} \sum_{j=1} ^{N_k}\nabla_{\bm \theta} \ell_k(\tau_k^j; \bm\theta)$ is an unbiased estimator of the true gradient $\bm g_k = \mathbb{E}_{\tau\sim \mathbb{P}_k}[\nabla_{\bm\theta} \ell_k(\tau; \bm\theta)]$, its variance scales inversely with the sample size:
$\mathrm{Var}[\hat{\bm g}_k]=\mathbb{E}[\|\hat{\bm g}_k-\bm g_k\|^2]=\mathcal{O}{(1/N_k)}$.
It indicates that limited data quantity inflates gradient variance.



\subsection{Data Sharing Strategy}
Cross-task data sharing is a common approach to increase effective sample sizes through relabeling, allowing trajectories from a source task $i$ to provide behavioral evidence for a target task $k$ \citep{kalashnikov2021mt,yu2022leverage,mitchell2021offline}.
This is particularly relevant in auto-bidding, where different tasks share identical transition and reward mechanisms despite varying constraints.
However, this approach inherently alters the underlying training distribution and poses distribution shifts \citep{yu2021conservative}. 

While existing offline RL methods utilize value-based signals to mitigate these shifts \citep{yu2021conservative,bai2024pessimistic}, they are largely incompatible with generative auto-bidding.
In generative scenarios, the objective shifts from optimizing a scalar value function to matching complex distributions of trajectories.
This misalignment presents an underexplored, technical challenge: 
Naive data sharing yields a low-fidelity approximation of the target trajectory distribution, causing the model to fail to satisfy the target constraints.
Consequently, achieving performance gains on data-scarce tasks requires a more sophisticated mechanism to navigate these distributional discrepancies.

\section{Method} \label{method_sec}
\begin{figure*}[t]
  \centering
\includegraphics[width=1.0\textwidth]{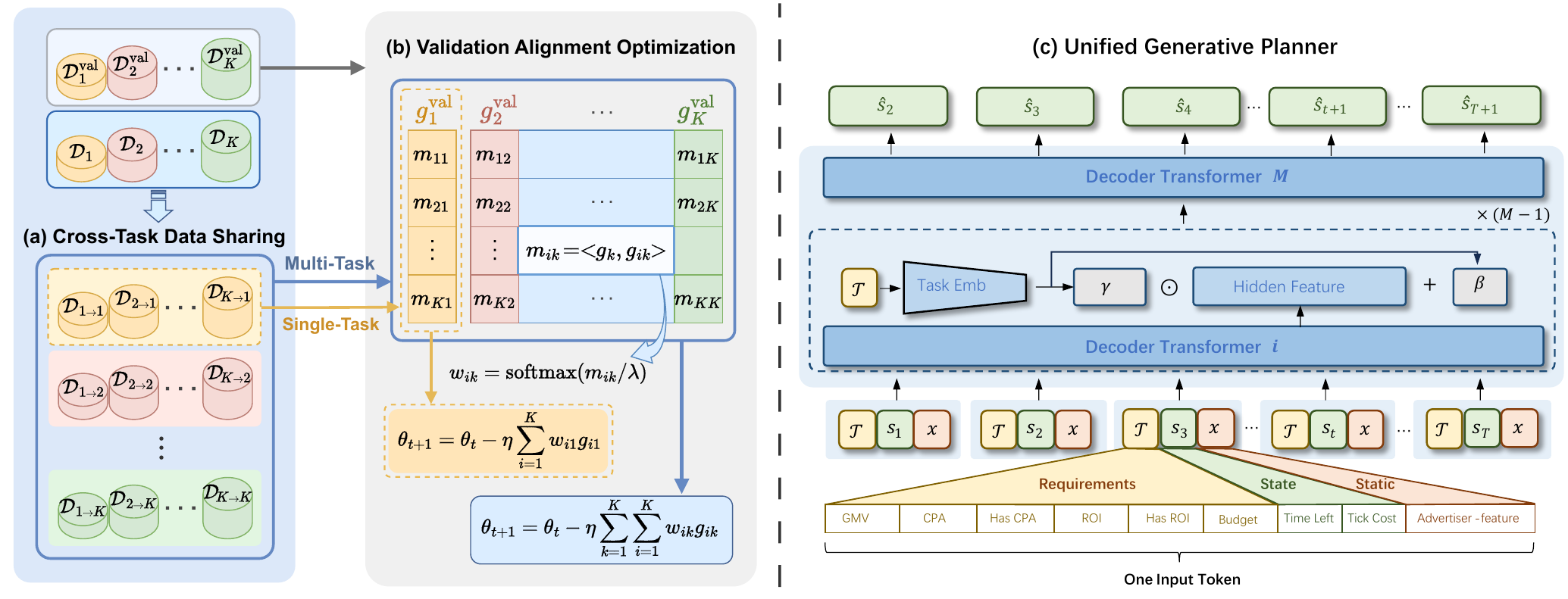}
  \caption{\textbf{Overview of the VAO framework and multi-task architectures.}
  (a) Data Sharing: Training data is shared across tasks via relabeling, with held-out validation sets drawn from the same task distributions.
  (b) VAO Optimization: Weight learning through alignment between target and cross-task gradients, supporting both single-task and multi-task learning.
  (c) Architecture: The unified generative planner architecture for multi-task learning.
  }
  \label{fig:main}
  \vspace{-5mm}
\end{figure*}

This section examines the inherent limitations of naive cross-task data sharing and presents validation-aligned optimization (VAO), a principled data-sharing strategy for generative auto-bidding to facilitate effective knowledge transfer.
We also provide rigorous theoretical guarantees for the performance improvements afforded by VAO.
For practical deployment, VAO is seamlessly integrated into a unified multi-task learning framework, enabling a single model instance to handle all tasks concurrently, as illustrated in Fig.~\ref{fig:main}.




\subsection{The Pitfalls of Naive Cross-Task Data Sharing}
A straightforward approach to cross-task data sharing is to pool trajectories from all tasks, relabel them with the target task's condition signal, and train on the shared dataset.
Let $\mathcal{D}_{i\rightarrow k}=\{(\tau_i,\bm y_k(\tau_i)) \mid (\tau_i,\bm y_i(\tau_i))\in\mathcal{D}_i\}$ denote the dataset from source task $i$ relabeled for target task $k$, and $\hat{\mathcal{D}}_k=\bigcup_{i=1}^K \mathcal{D}_{i \rightarrow k}$ denotes the shared dataset.
Under naive sharing, the empirical training distribution is a mixture with weights $\alpha_i\triangleq\frac{N_i}{N}$ proportional to each task’s data quantity, where $N\triangleq\sum_{k=1}^K N_k$ denotes the total sample size.
Define the empirical risk on dataset $\mathcal{D}_i$ but evaluated under the target loss $\ell_k$ as:
$\hat{\mathcal{R}}_i(\bm{\theta}) = \frac{1}{N_i}\sum_{(\tau_i,\bm y_k(\tau_i))\in\mathcal{D}_{i \rightarrow k}}\ell_k(\tau_i; \bm{\theta})$.
The naive training objective is then the weighted mixture: $\hat{\mathcal{R}}^{\mathrm{Naive}}_k(\bm{\theta}) = \sum_{i=1}^K \alpha_i \hat{\mathcal{R}}_i(\bm{\theta})
= \frac{1}{N}\sum_{i=1}^K\sum_{(\tau_i,\bm y_k(\tau_i))\in\mathcal{D}_{i \rightarrow k}}\ell_k(\tau_i; \bm{\theta})$.

The naive data sharing strategy increases available examples, reduces gradient variance, and appears to mitigate data scarcity, especially for minority tasks.
However, it hardly handles the generalization issue caused by distribution shifts across tasks.
Discrepancies in bidding constraints across tasks manifest as distribution shifts in trajectory space, systematically degrading target-task performance.

\textbf{Theoretical Pitfalls Analysis.}
To rigorously analyze this degradation, we characterize the generalization gap between the empirical naive objective $\hat{\mathcal{R}}_{k}^\mathrm{Naive
}(\bm{\theta}) $ and the true target risk $ \mathcal{R}_k(\bm{\theta}) = \mathbb{E}_{(\tau,\bm y(\tau)) \sim \mathbb{P}_k}[\ell_k(\tau; \bm{\theta})]$ in 
Theorem \ref{thm:naive_bound}.

\begin{assumption}\label{assump:regularity} 
We assume: (i) the loss function $\ell_k(\tau; \bm{\theta})$ is bounded, i.e., $\ell_k(\tau; \bm{\theta})\in [0,M]$ for all $\tau, \bm{\theta}$; (ii) $\ell_k(\cdot; \bm{\theta})$ is $L$-Lipschitz with respect to $\bm{\theta}$; (iii) the hypothesis class $\Theta$ has finite Rademacher complexity $\mathfrak{R}_N(\Theta)$ \citep{shalev2014understanding}. 
\end{assumption}

\begin{theorem}[Generalization Bound under Naive Cross-Task Data Sharing]
\label{thm:naive_bound}
Under Assumption \ref{assump:regularity}, with probability at least $1-\delta$, we have the following generalization bound in the presence of naive cross-task data sharing:
\begin{equation}
\begin{aligned}
    \mathcal{R}_k(\bm{\theta}) &\leq \hat{\mathcal{R}}_{k}^\mathrm{Naive
}(\bm{\theta}) 
+  \underbrace{M \sum_{i \neq k} \colorbox{cyan!20}{$\myheight \alpha_i$} \colorbox{orange!20}{$\myheight d_{\mathrm{TV}}(\mathbb{P}_i, \mathbb{P}_k)$}}_{\text{distribution shift bias}}\\
&+  \underbrace{2L \, \mathfrak{R}_{N}(\Theta) 
+ M \sqrt{\frac{\log(2/\delta)}{2N}}}_{\text{estimation error}},
\end{aligned}
\end{equation}
where $d_{\mathrm{TV}}(\mathbb{P}_i, \mathbb{P}_k)\triangleq\sup_A|\mathbb{P}_i(A)-\mathbb{P}_k(A)|$ denotes the total variation distance between task distributions.
\end{theorem}

The proof of Theorem \ref{thm:naive_bound} is given in Appendix \ref{proof:naive_generalization_bound}, and the reasonability of mild Assumption \ref{assump:regularity} is explained in Appendix \ref{app:reasonablity of assumption}.
Theorem \ref{thm:naive_bound} reveals a fundamental gap of naive sharing from the distribution shift term.
This term comprises two factors, including the data-proportional weights $\alpha_i$ and the divergence between the distribution of trajectories in task $i$ and the target distribution {$d_{\mathrm{TV}}(\mathbb{P}_i, \mathbb{P}_k)$.
This bias accumulates over all source tasks $i \neq k$, where each task's contribution scales with its portion and extent of divergence.

For data-scarce tasks where $\alpha_k \ll \sum_{i \neq k} \alpha_i$, the distribution shift term becomes the dominant factor in the generalization error bound. This remains true even for large total sample sizes $N$, as data-proportional weighting $\{\alpha_i\}$ fails to account for the alignment of each source with the target.
Consequently, a data-abundant task $i$ with a large $\alpha_i$ but a high $d_{\mathrm{TV}}(\mathbb{P}_i, \mathbb{P}_k)$ introduces significant bias, while well-aligned minority tasks with low $d_{\mathrm{TV}}$ are underutilized despite their potential benefit.


\textbf{Distribution Shift Impact Analysis.}
We further investigate how performance degradation from distribution shift unfolds during the naive data-sharing process.
Specifically, naive data-sharing incorporates an expected gradient derived from trajectories of the source distribution $\mathbb{P}_i$ but evaluated under the target loss $l_k$, i.e., ${\bm g}_{ik}\triangleq\mathbb{E}_{(\tau,\bm y(\tau) )\sim \mathbb{P}_i}[\nabla_{\bm\theta} \ell_k(\tau; \bm\theta)]$.
We refer to this gradient as the \textit{cross-task gradient}.
Note that the expected gradient of the empirical naive objective is a weighted sum of the cross-task gradients, i.e., 
\begin{equation}\label{eq:naive_gradient}
    \mathbb{E}[\nabla_{\bm{\theta}} \hat{\mathcal{R}}^{\mathrm{Naive}}_k(\bm{\theta})] =  \sum_{i=1}^K \alpha_i\mathbb{E}[\nabla_{\bm{\theta}} \hat{\mathcal{R}}_i(\bm{\theta})]=\sum_{i=1}^K  \alpha_i{\bm g}_{ik},
\end{equation}
whereas the true gradient for task $k$ is $\bm{g}_k=\bm{g}_{kk}$.
The gradient bias of naive sharing is thereby given by:
\begin{equation}\label{eq:gradient_bias}
    \mathrm{Bias}_k =\| \sum_{i=1}^K \alpha_i \bm g_{ik} - \bm g_k\| = \| \sum_{i \neq k} \colorbox{cyan!20}{$\myheight \alpha_i$} \colorbox{orange!20}{$\myheight (\bm g_{ik} - \bm g_k)$}\|,
\end{equation}
where we leverage the property of $\sum_{i=1}^K\alpha_i=1$.

It can be observed from Eq. \eqref{eq:gradient_bias} that the gradient estimation bias originates exclusively from the cross-task terms where $i\ne k$. Each source task's contribution is jointly determined by its data proportion $\alpha _{i}$ and its gradient mismatch $(\bm {g}_{ik}-\bm{g}_{k})$. This decomposition directly mirrors the two components of the distribution shift term established in Theorem~\ref{thm:naive_bound}.
Particularly, the gradient mismatch $(\bm g_{ik} - \bm g_k)$ reflects the distributional divergence $d_{\mathrm{TV}}(\mathbb{P}_i, \mathbb{P}_k)$.
Heterogeneous bidding constraints cause gradients from $\mathbb{P}_i$ and $\mathbb{P}_k$ to diverge in direction.
For data-scarce tasks with small $\alpha_k$, the gradient bias is dominated by majority tasks ($i\neq k$), potentially steering optimization away from the target optimum and eventually converging to a suboptimal solution $\bm\theta^{\mathrm{Naive}}$, which performs poorly on the target distribution.

The preceding analysis reveals that data-proportional weighting \(\{\alpha _{i}\}\) is intrinsically suboptimal for target task performance. This underscores the need to identify task-specific weights that better align source-task contributions with target objectives. However, directly estimating distributional divergences and gradient mismatches remains computationally prohibitive in offline settings, as it requires either full knowledge of trajectory distributions or exhaustive computation of all cross-task gradient interactions.
To this end, we present a principled approach to cross-task data-sharing in the next section.

\subsection{Validation Aligned Optimization Strategy}
This section presents a principled approach to cross-task data sharing that directly optimizes for generalization to the target task.
Our basic idea is to use a held-out validation set from the target task to learn source-task weights that enhance target performance, accounting for distributional divergence and gradient alignment.

\textbf{Validation-Aligned Adaptive Weighting.}
Based on the results established in Theorem \ref{thm:naive_bound}, source tasks contribute heterogeneously to target performance, with their influence being critically dependent on their distributional divergence.
Instead of using data-proportional weights $\{\alpha_i\}$ that overlook this heterogeneity, we turn to design adaptive task-specific weights $\{w_{ik}\}$ for each target task $k$, where $\sum_{i=1}^K w_{ik}=1$.
These weights are optimized using a held-out validation set $\mathcal{D}_k^{\mathrm{val}}$ from task $k$, which yields an unbiased empirical estimate of the true target risk $\mathcal{R}_k(\bm{\theta})$.
This approach enables us to directly optimize for generalization performance, thereby circumventing the need for intractable estimates of distributional divergences or gradient mismatches. By selecting task weights that minimize validation risk, we implicitly align the training dynamics with the target's generalization objectives.
 
Starting from parameters ${\bm\theta}_t$ at the $t$-th iteration, a gradient update with step size $\eta>0$ produces $
    {\bm\theta}_{t+1}
= {\bm\theta}_t - \eta \sum_{i=1}^K w_{ik}\,{\bm g}_{ik}$.
Applying a first-order Taylor expansion of the validation risk $\mathcal{R}_k^{\mathrm{val}}({\bm\theta})$ around ${\bm\theta}_t$ yields:
\begin{equation}
\begin{aligned}
    \mathcal{R}_k^{\mathrm{val}}({\bm\theta}_{t+1})  =& \mathcal{R}_k^{\mathrm{val}}({\bm\theta}_t) - \eta \Big\langle \nabla_{\bm \theta}\mathcal{R}_k^{\mathrm{val}}(\bm\theta_t), \; \sum_{i=1}^K w_{ik} \bm g_{ik} \Big\rangle \\ &+ \mathcal{O}(||{\bm\theta}_{t+1}-{\bm\theta}_t||^2),\notag
\end{aligned}
\end{equation}
where $\nabla_{\bm \theta}\mathcal{R}_k^{\mathrm{val}}(\bm\theta_t)=\bm g_k$.
The resulting \textit{change} in validation loss is then approximated by:
\begin{equation}
\Delta \mathcal{R}_k^{\mathrm{val}} \approx -\eta \sum_{i=1}^K \colorbox{cyan!20}{$w_{ik}$} \colorbox{orange!20}{$m_{ik}$}, \quad \colorbox{orange!20}{$m_{ik}\triangleq\langle{\bm{g}}_k,{\bm{g}}_{ik}\rangle$}.
\label{eq:val_loss_change}
\end{equation}
Eq. \eqref{eq:val_loss_change} connects validation-based weighting to the theoretical insights from Theorem \ref{thm:naive_bound}.
The adaptive weight {$w_{ik}$} replaces the data-proportional $\alpha_i$, enabling us to down-weight sources with large data size but poor alignment.
The gradient alignment {$m_{ik}=\langle{\bm{g}}_k,{\bm{g}}_{ik}\rangle$} implicitly captures both distributional divergence and gradient mismatch, guiding the optimization toward beneficial sources ($m_{ik} > 0$) and away from harmful ones ($m_{ik} < 0$).
By optimizing $\{w_{ik}\}$ to minimize validation risk, we obtain a practical weighting scheme that adapts to target generalization without explicitly estimating intractable divergences.

\textbf{Entropy-Regularized Weight Optimization.}
Equation \eqref{eq:val_loss_change} measures how effectively each data source drives the policy toward the target task. 
However, maximizing $\sum_{i=1}^K w_{ik}m_{ik}$ tends to lead to a sparse assignment, favoring only the task with the highest $m_{ik}$. 
This approach overlooks the advantages of a diverse set of auxiliary tasks. 
To promote a more balanced integration of cross-task knowledge, we propose an entropy-regularized objective:
\begin{equation}\label{eq:regularized_obj}
\max_{\mathbf{w}_k \in \Delta^{K-1}} \sum_{i=1}^K w_{ik} m_{ik}-\lambda \sum_{i=1}^K w_{ik} \ln w_{ik},
\end{equation}
where $\lambda>0$ controls the strength of regularization.
We rigorously demonstrate that such a regularized convex programming problem has a closed-form solution:
\begin{equation}
w_{ik}^* = \frac{\exp(m_{ik}/\lambda)}{\sum_{j=1}^K \exp(m_{jk}/\lambda)}.
\end{equation}

\begin{remark}
The optimal weights satisfy $w_{ik}^* =\operatorname{softmax}\!\left(m_{ik}/\lambda\right)$ with $\lambda$ as a temperature hyperparameter. 
As $\lambda \to 0$, the weight allocation becomes hard; as $\lambda \to \infty$, the weights approach uniform.
Entropy regularization thus provides a principled mechanism for interpolating between these two patterns, ensuring that the target task benefits from a diverse ensemble of auxiliary sources.
\end{remark}

Notably, we establish a theoretical guarantee that our proposed method outperforms the naive data-sharing approach in Theorem \ref{thm:improvement_guaranteen}. The proof is given in Appendix \ref{proof:thm:improvement_guaranteen}.

\begin{theorem}[Improvement Guarantee of VAO]\label{thm:improvement_guaranteen}
Under the same setting, let $\bm{\theta}_{t+1}^{\mathrm{VAO}}$ and $\bm{\theta}_{t+1}^{\mathrm{Naive}}$ denote the parameters updated after one gradient step from $\bm \theta_t$ with VAO weights $\{w_{ik}^*\}$ and naive weights $\{\alpha_i\}$, respectively. Then, we have
\begin{equation}
\begin{aligned}
    \mathcal{R}_k(\bm{\theta}_{t+1}^{\mathrm{VAO}}) & \leq \mathcal{R}_k(\bm{\theta}_{t+1}^{\mathrm{Naive}})
     - \eta \Delta + \mathcal{O}(\eta^2), \\
     \Delta &=\sum_{i=1}^K (w_{ik}^* - \alpha_i) m_{ik},
\end{aligned}
\end{equation}
where $\Delta \geq 0$, and $\Delta > 0$ whenever the naive weights $\{\alpha_i\}$ deviate from the alignment-optimized weights $\{w_{ik}^*\}$. 
\end{theorem}

Theorem \ref{thm:improvement_guaranteen} shows that VAO mitigates the distribution-shift bias identified in Theorem \ref{thm:naive_bound}.
While Theorem \ref{thm:naive_bound} shows that naive sharing is limited by distributional divergence, VAO operationalizes this via gradient alignment, replacing static weights $\alpha_i$ with adaptive weights $w_{ik}^*$ and prioritizing sources to best improve target performance.
The positive improvement $\Delta$ ensures VAO consistently steers the model toward lower generalization error, tightening the theoretical bound that naive sharing fails to achieve.

\begin{algorithm}[t]
{\small 
    \caption{Validation Aligned Optimization (VAO)}
    \begin{algorithmic}[1]
        \STATE \textbf{Input}:  
        Maximum iteration number $T$;
        Learning rate $\eta$; 
        Temperature hyperparameter $\lambda$;
        Target task $k$;
        Validation set $\mathcal{D}_k^{\mathrm{val}}$; 
        
        \STATE Initialize model parameters ${\bm\theta}_0$; 
        \STATE Relabel source task data to target task data $\{\mathcal{D}_{i \rightarrow k}\}_{i=1}^K$;
    
        \FOR{$t = 0$ to $T$}
            \STATE Compute validation gradient $\bm g_k$ through $\mathcal{D}_k^{\mathrm{val}}$;
            \FOR{$i = 0$ to $K$}
                \STATE Compute cross-task training gradient $\bm g_{ik}$ through $\mathcal{D}_{i \rightarrow k}$;
                
                \STATE Compute marginal gains $m_{ik} = \langle \bm g_k, \bm g_{ik}\rangle$;
                
                \STATE Compute weights $w_{ik} = \frac{\exp(m_{ik}/\lambda)}{\sum_{j=1}^K \exp(m_{jk}/\lambda)}$;
            \ENDFOR
    
            \STATE Update parameters ${\bm\theta}_{t+1} = {\bm\theta}_t - \eta \sum_i w_{ik} \bm g_{ik}$;

            
        \ENDFOR
        
    \end{algorithmic}
    }
\label{algorithm}
\end{algorithm}

\subsection{Practical Algorithm Design}
In real-world systems, maintaining a separate model for each bidding task incurs substantial overhead and fails to leverage shared structures across tasks.
Fortunately, our validation-aligned optimization scheme can be seamlessly extended to a multi-task setting, where all tasks jointly update a single model.
The resulting update rule is:
\begin{equation}
    {\bm\theta}_{t+1}
= {\bm\theta}_t - \eta \sum_{k=1}^K\sum_{i=1}^K w_{ik}\,{\bm g}_{ik}.
\end{equation}
This enables a single model to handle multiple bidding tasks simultaneously while preserving task‑specific performance through adaptive gradient alignment.

\textbf{Model Architecture.}
Here, we adopt a shared backbone to extract generalizable representations across tasks, improving parameter efficiency and enabling knowledge transfer.
Specifically, as shown in Fig. \ref{fig:main}, we employ a Causal Transformer with multiple Transformer blocks as the shared backbone to generate trajectories in a task-agnostic manner. 
Each input token is constructed by concatenating the trajectory requirements, the state $\bm {s}_{t}$, and the advertiser's static features. Here, the \textit{requirement} term encompasses all specific constraints and objectives across multiple tasks. 
To enhance the model's sensitivity to these requirements, we incorporate a Feature-wise Linear Modulation (FiLM) mechanism \cite{perez2018film} into the architecture.
Specifically, the FiLM module integrates a residual structure into the model's hidden layers. It applies a linear transformation to the hidden features, conditioned on the requirements provided in the input tokens, which can be expressed as: 
\begin{equation}
    \text{FiLM}(\text{Hidden Feature})=\gamma(\bm y)\cdot \text{Hidden Feature}+\beta(\bm y)\notag
\end{equation}
where $\bm y$ denotes the embedding of the requirements. 
The FiLM output is processed by the next Transformer block. 
During online inference, irrelevant requirements are masked to keep the model focused on task-specific constraints.

\section{Experiments} \label{exp_sec}


\setlength{\tabcolsep}{9.0pt}
\begin{table*}[t]
\caption{\textbf{Performance comparison with baselines on three bidding tasks under single-task learning settings.} Mean and standard error are reported across five seeds. Metrics include value and score for each task. 
\textbf{Bold} denotes the best results.
$\uparrow$ denotes the higher the better.}
\small
\centering
\begin{tabular}{ll|cccccc}
\toprule
 \multicolumn{2}{c|}{\multirow{2}{*}{\textbf{Method}}} & \multicolumn{2}{c}{\textbf{CPA-MCB}}  & \multicolumn{2}{c}{\textbf{TargetROAS}}  & \multicolumn{2}{c}{\textbf{BCB}} \\
\cmidrule(lr){3-4}\cmidrule(lr){5-6}\cmidrule(lr){7-8}
\multicolumn{2}{c|}{~} & Value $\uparrow$ & Score $\uparrow$ & Value $\uparrow$ & Score $\uparrow$ & Value $\uparrow$ & Score $\uparrow$\\
\midrule
MDMM & No Sharing & 25.46\scriptsize{$\pm$0.54} & 19.14\scriptsize{$\pm$0.55} & 24.05\scriptsize{$\pm$0.36}& 24.32\scriptsize{$\pm$0.43} & 19.37\scriptsize{$\pm$0.25} & 13.34\scriptsize{$\pm$0.26} \\
\midrule
\multirow{2}{*}{ODMM} & Naive Sharing & 22.63\scriptsize{$\pm$1.25} & 17.57\scriptsize{$\pm$0.92} &28.01\scriptsize{$\pm$0.33}  & 28.41\scriptsize{$\pm$0.37} &20.52\scriptsize{$\pm$0.25} & 13.97\scriptsize{$\pm$0.23}  \\
 & \cellcolor{myhighlight}\textbf{VAO (Ours)} &\cellcolor{myhighlight}\textbf{34.68\scriptsize{$\pm$0.25}} &
\cellcolor{myhighlight}\textbf{29.24\scriptsize{$\pm$0.17}} & 
\cellcolor{myhighlight}\textbf{32.80\scriptsize{$\pm$0.07}} & 
\cellcolor{myhighlight}\textbf{33.42\scriptsize{$\pm$0.11}} & \cellcolor{myhighlight}\textbf{28.13\scriptsize{$\pm$0.13}} & \cellcolor{myhighlight}\textbf{24.27\scriptsize{$\pm$0.12}} 
\\
\bottomrule
\end{tabular}
\label{table_single_task}
\end{table*}

\setlength{\tabcolsep}{4.8pt}
\begin{table*}[t]
\caption{\textbf{Performance comparison with baselines on three bidding tasks under multi-task learning settings.} Mean and standard error are reported across five seeds. Metrics include value and score for each task and overall MTL performance $\Delta m\%$. \textbf{Bold} and \underline{underlined} denote the best and the most competitive results. $\downarrow$ denotes the lower the better.}
\small
\centering
\begin{tabular}{ll|cccccc|c}
\toprule
\multicolumn{2}{c|}{\multirow{2}{*}{\textbf{Method}}} & \multicolumn{2}{c}{\textbf{CPA-MCB}}  & \multicolumn{2}{c}{\textbf{TargetROAS}}  & \multicolumn{2}{c|}{\textbf{BCB}} & \multirow{2}{*}{$\bm \Delta \bm m\% \downarrow$}\\
\cmidrule(lr){3-4}\cmidrule(lr){5-6}\cmidrule(lr){7-8}
\multicolumn{2}{c|}{~} & Value $\uparrow$ & Score $\uparrow$ & Value $\uparrow$ & Score $\uparrow$ & Value $\uparrow$ & Score $\uparrow$\\
\midrule
MDMM & STL (No Sharing) &25.46\scriptsize{$\pm$0.54} & 19.14\scriptsize{$\pm$0.55} & 24.05\scriptsize{$\pm$0.36}& 24.32\scriptsize{$\pm$0.43} & 19.37\scriptsize{$\pm$0.25} & 13.34\scriptsize{$\pm$0.26} & - \\
\midrule
\multirow{4}{*}{MDOM} & EW & 21.12\scriptsize{$\pm$1.39} & 19.26\scriptsize{$\pm$1.27} & 15.43\scriptsize{$\pm$0.56} & 13.38\scriptsize{$\pm$0.64}& 12.93\scriptsize{$\pm$0.74} & 7.22\scriptsize{$\pm$0.55} & 29.39 \\
& FAMO \citep{liu2023famo} & 23.00\scriptsize{$\pm$0.16} & 18.80\scriptsize{$\pm$0.20} & 20.94\scriptsize{$\pm$0.28} & 20.19\scriptsize{$\pm$0.29}& 17.99\scriptsize{$\pm$0.26} & 12.21\scriptsize{$\pm$0.27} & 9.49 \\
& PCGrad \citep{yu2020gradient} & 31.65\scriptsize{$\pm$0.23} & \underline{27.87\scriptsize{$\pm$0.25}}& 25.60\scriptsize{$\pm$0.41}  & 24.56\scriptsize{$\pm$0.51}& 21.63\scriptsize{$\pm$0.18} & 14.72\scriptsize{$\pm$0.16}& -16.56 \\
& FairGrad \citep{ban2024fair} & 20.35\scriptsize{$\pm$0.57} & 15.40\scriptsize{$\pm$0.38}& 22.73\scriptsize{$\pm$0.30} & 22.45\scriptsize{$\pm$0.37}& 19.81\scriptsize{$\pm$0.17} & 14.82\scriptsize{$\pm$0.19}& 6.57 \\
\midrule
\multirow{5}{*}{ODOM} & EW & 27.45\scriptsize{$\pm$0.41} & 21.95\scriptsize{$\pm$0.44}& 27.58\scriptsize{$\pm$0.36} & 27.57\scriptsize{$\pm$0.39}& 26.34\scriptsize{$\pm$0.41} & 21.32\scriptsize{$\pm$0.48}& -24.39 \\
& FAMO \citep{liu2023famo} & 26.90\scriptsize{$\pm$0.27} & 20.28\scriptsize{$\pm$0.35}& 29.43\scriptsize{$\pm$0.18} & \underline{30.03\scriptsize{$\pm$0.23}}& \underline{27.29\scriptsize{$\pm$0.13}} &\underline{23.85\scriptsize{$\pm$0.15}} & -29.52 \\
& PCGrad \citep{yu2020gradient} & 30.33\scriptsize{$\pm$1.01} & 25.08\scriptsize{$\pm$1.01} & \underline{29.46\scriptsize{$\pm$0.61}} & 29.84\scriptsize{$\pm$0.65}& 26.17\scriptsize{$\pm$0.62} &20.51\scriptsize{$\pm$0.69} & -30.70 \\
& FairGrad \citep{ban2024fair} & \underline{33.64\scriptsize{$\pm$0.39}} & 24.50\scriptsize{$\pm$0.36}& 27.71\scriptsize{$\pm$0.26} &27.90\scriptsize{$\pm$0.22} & 25.77\scriptsize{$\pm$0.16} & 22.73\scriptsize{$\pm$0.15} & \underline{-32.25}  \\
& \cellcolor{myhighlight}  \textbf{VAO (Ours)} &\cellcolor{myhighlight}\textbf{36.10\scriptsize{$\pm$0.25}} &
\cellcolor{myhighlight}\textbf{31.48\scriptsize{$\pm$0.13}} & \cellcolor{myhighlight}\textbf{30.38\scriptsize{$\pm$0.15}} & \cellcolor{myhighlight}\textbf{30.94\scriptsize{$\pm$0.14}} & \cellcolor{myhighlight}\textbf{28.01\scriptsize{$\pm$0.13}} & \cellcolor{myhighlight}\textbf{23.97\scriptsize{$\pm$0.19}}& \cellcolor{myhighlight}\textbf{-47.35}\\
\bottomrule
\end{tabular}
\label{table_multi_task}
\vspace{-2mm}
\end{table*}

\subsection{Experimental Setup}

\textbf{Benchmark and Environment.}
We conduct experiments in AuctionNet \citep{su2024auctionnet}, a large-scale ad auction benchmark derived from a real-world online advertising platform of Alibaba \citep{xu2024auto}.
We collect three bidding task datasets comprising multi-episode trajectories of competing agents.
Each episode corresponds to a one-day 
advertising delivery period and is divided into 48 time intervals of 30 minutes each, where agents observe states and determine bids sequentially.
To mimic real-world advertiser preferences, we assign each task to a disjoint set of advertisers, resulting in imbalanced sample sizes across tasks.
This data proportion reflects real-world advertising platform statistics, as detailed in Table \ref{tab:data_split}.
Validation and test sets preserve these task-specific scales to ensure a realistic evaluation.

\textbf{Bidding Tasks.} Our experiments focus on three distinct bidding tasks, all sharing the common objective of maximizing total GMV but subject to different constraints.
\textbf{CPA-MCB} ensures the realized cost-per-acquisition (CPA) and total cost stay below predefined CPA and budget constraints by the end of the episode;
\textbf{TargetROAS} aims to achieve a target ROAS while respecting a budget constraint;
\textbf{BCB} focuses solely on adhering to a strict budget constraint.


\textbf{Baselines.}
To evaluate the effectiveness of VAO, we compare it against baselines categorized by their data utilization and model architecture across two settings: single-task and multi-task.
Single-task baselines include: (i) \textbf{MDMM} (Multi-Data Multi-Model), where independent models are trained on task-specific datasets; and (ii) \textbf{ODMM} (One-Data Multi-Model), where separate models are trained on a pooled dataset.
Multi-task baselines utilize a unified model architecture under two data configurations: (i) \textbf{MDOM} (Multi-Data One-Model), the standard multi-task learning setup where each task objective is optimized using its respective dataset; (ii) \textbf{ODOM} (One-Data One-Model), an extension of naive sharing to the multi-task setting, where a shared model is trained on the pooled dataset.
For both MDOM and ODOM, we compare several optimization strategies: EW (Equal Weighting), which assigns uniform loss weights; FAMO \citep{liu2023famo}, which balances task losses by ensuring each task's loss decreases approximately at an equal rate; PCGrad \citep{yu2020gradient}, which projects conflicting gradients to mitigate interference; and FairGrad \citep{ban2024fair}, which adjusts gradients through fair resource allocation to ensure balanced task updates.



\textbf{Evaluations.}
To provide a comprehensive assessment, we adopt two evaluation metrics:
Value, defined as the total conversion value accumulated over the bidding period; and Score, calculated by multiplying the Value by a penalty term that reflects the degree of satisfaction of predefined constraints (e.g., CPA, TargetROI, and budget).
We also adopt a common metric for evaluating multi-task learning performance: average relative performance drop, denoted by $\Delta m\%$ \citep{liu2023famo,shen2024go4align}. 
It quantifies the average degradation of an MTL method compared to the STL baseline across all tasks. Specifically, $\Delta m\%=\frac{1}{S}\sum_{s=1}^S-(M_{s}-B_{s})/B_{s}\times 100$, where $M_{s}$ and $B_{s}$ are the $s$-th metric values for the MTL method and STL baseline, respectively, and $S$ is the total number of metrics for all tasks.
Lower $\Delta m\%$ indicates better MTL performance, and negative values signify MTL outperforms STL.



\setlength{\tabcolsep}{9.0pt}
\begin{table*}
\caption{\textbf{Ablation study on FiLM modules.} Performance comparison without and with FiLM in the multi-task architecture. Mean and standard error are reported over five runs.}
\small
\centering
\begin{tabular}{l|ccccccc}
\toprule
 {\multirow{2}{*}{\textbf{Method}}} & \multicolumn{2}{c}{\textbf{CPA-MCB}}  & \multicolumn{2}{c}{\textbf{TargetROAS}}  & \multicolumn{2}{c}{\textbf{BCB}} & \multirow{2}{*}{$\bm \Delta \bm m\% \downarrow$} \\
\cmidrule(lr){2-3}\cmidrule(lr){4-5}\cmidrule(lr){6-7}
& Value $\uparrow$ & Score $\uparrow$ & Value $\uparrow$ & Score $\uparrow$ & Value $\uparrow$ & Score $\uparrow$ & \\
\midrule
 w/o FiLM & 32.73\scriptsize{$\pm$0.12} & 29.30\scriptsize{$\pm$0.22} & 29.77\scriptsize{$\pm$0.13}& 29.70\scriptsize{$\pm$0.15} & 27.35\scriptsize{$\pm$0.44} & 23.06\scriptsize{$\pm$0.52} & -40.27 \\
 \textbf{VAO (Ours)} & \textbf{36.10\scriptsize{$\pm$0.25}} &
\textbf{31.48\scriptsize{$\pm$0.13}} & \textbf{30.38\scriptsize{$\pm$0.15}} & \textbf{30.94\scriptsize{$\pm$0.14}} & \textbf{28.01\scriptsize{$\pm$0.13}} & \textbf{23.97\scriptsize{$\pm$0.19}}& \textbf{-47.35}
\\
\bottomrule
\end{tabular}
\label{table_ablation_film}
\vspace{-3mm}
\end{table*}


\subsection{Empirical Result Analysis}
We present performance comparisons for single-task settings in Table \ref{table_single_task}. 
In data-scarce tasks like CPA-MCB, naive sharing strategies degrade performance due to distribution shifts and provide minimal gains. 
In contrast, VAO significantly improves both Value and Score by adaptively weighting cross-task data based on validation alignment, ensuring training updates align with the target task’s generalization and preventing dominance by other tasks.

Table \ref{table_multi_task} presents performance comparisons across three bidding tasks under multi-task learning settings, where VAO substantially outperforms all baselines.
Our method achieves $\Delta m\% = -47.35$, representing a 46.8\% relative improvement over the best baseline of FairGrad, which achieves -32.25.
Notably, VAO excels particularly on the minority task CPA-MCB, validating that validation-aligned weighting effectively mitigates distribution shift by prioritizing updates that align with target generalization.
Unlike standard MTL methods that focus on balancing joint performance across all tasks, VAO employs task-specific adaptive weights to optimize each task individually by learning which auxiliary sources are truly beneficial.

Comparing MDOM with separate ODOM data versus pooled ODOM data, we observe heterogeneous effects across tasks under naive sharing.
For instance, under EW, CPA-MCB gains only 7.8\% in Value from 25.46 to 27.45, which is substantially less than the 36.0\% improvement seen on BCB from 19.37 to 26.34.
This heterogeneity reflects the distribution shift  identified in Theorem \ref{thm:naive_bound}: tasks benefit from pooling only when the auxiliary data is well aligned.
Beyond superior performance, VAO exhibits consistently lower variance and a smaller standard error than the baselines, indicating stable optimization.
This stability stems from validation-guided weighting, which suppresses contributions from misaligned sources and avoids gradient bias.
Crucially, this stability suggests that even a small validation set suffices for reliable learning of weights.

\subsection{Ablation Study}
\begin{figure}
    \begin{center}
  \includegraphics[width=0.47
\textwidth]{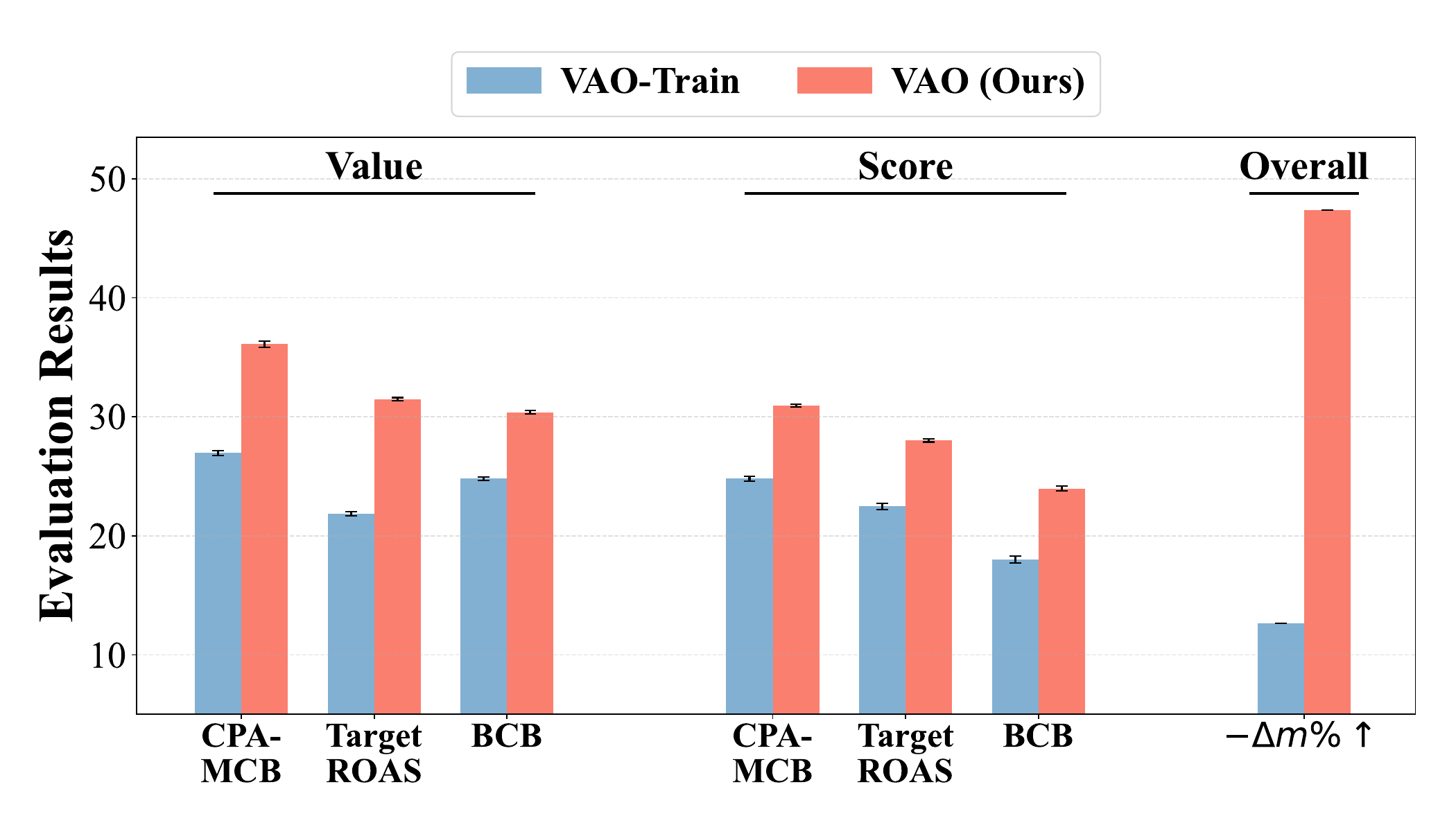}
  \end{center}
  \vspace{-3pt}
  \caption{\textbf{Ablation study on validation signal.} Comparison of VAO using validation data for alignment against VAO-Train using training data only. Error bars indicate standard error across five random seeds.
  }
  \label{fig:validation}
  \vspace{-20pt}
  \end{figure}

\textbf{Effects of Validation Signal.}
To evaluate the efficacy of the validation signal, we conduct an ablation study by using the training gradient as the alignment target instead of the held-out validation gradient.
This variant, denoted as ``VAO-Train'', aligns updates with an independent batch of training data rather than a validation set.
As shown in Fig. \ref{fig:validation}, VAO-Train consistently underperforms our validation-aligned method across all metrics.
This performance gap arises because using the training gradient as both the optimization direction and the alignment reference introduces self-referential bias.
The alignment target contains task-specific sampling noise that is highly correlated with the update, which risks overfitting to the training distribution.
In contrast, VAO leverages an external, unbiased signal from a held-out validation set, yielding a more reliable estimate of each source task’s true impact on target generalization.

\begin{figure}
    \begin{center}
  \includegraphics[width=0.47
\textwidth]{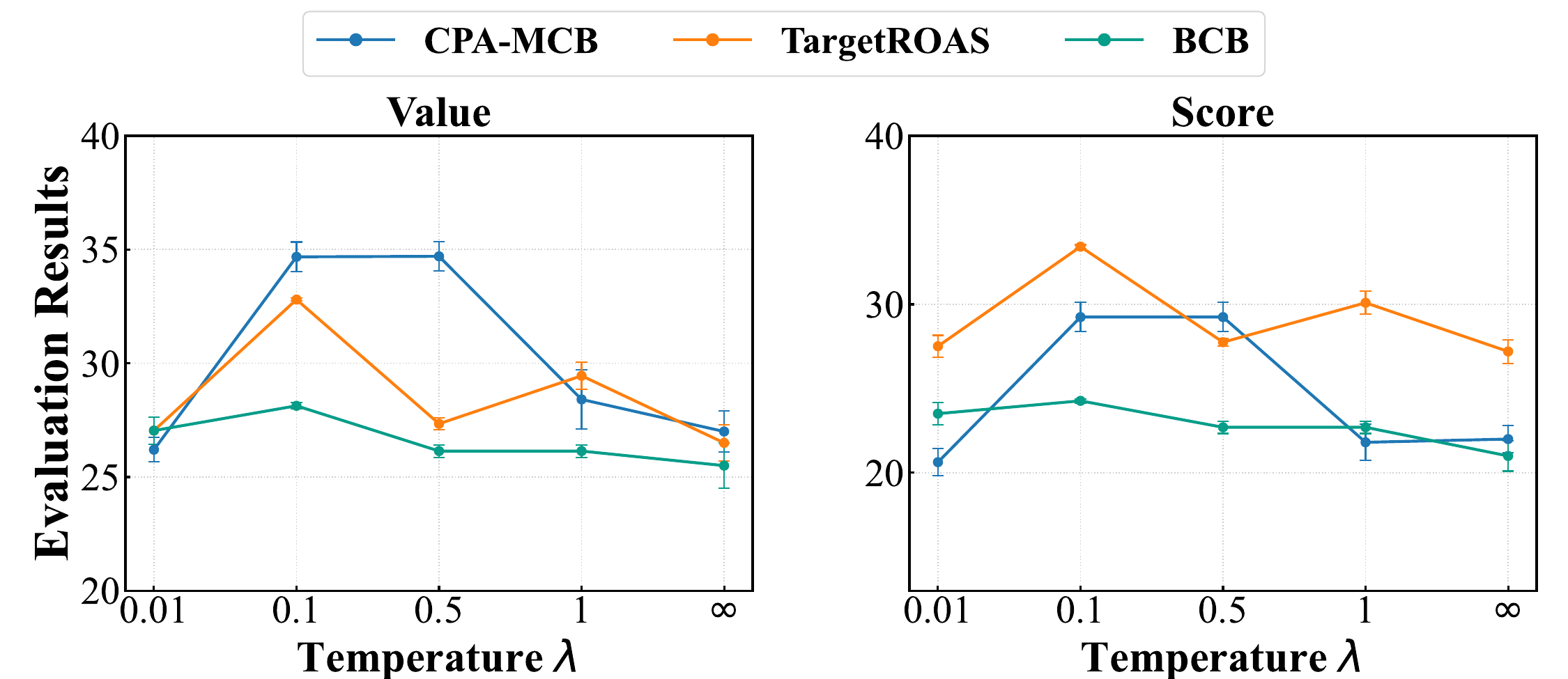}
  \end{center}
  \vspace{-3pt}
  \caption{\textbf{Ablation study on the hyperparameter $\lambda$.} Error bars denote standard error across five random seeds.
  }
  \label{fig:lambda}
  \vspace{-20pt}
  \end{figure}


\textbf{Influence of Temperature $\lambda$.}
To investigate the role of entropy regularization, we ablate the temperature parameter $\lambda$, which controls the strength of regularization.
As $\lambda \to 0$, the weights collapse to a hard assignment on the single most aligned task.
As shown in Fig. \ref{fig:lambda}, at $\lambda =0.01$, this causes poor generalization, ignoring potentially helpful signals from other sources.
As $\lambda \to \infty$, weights become uniform, treating all tasks equally and reintroducing distribution shift.
Moderate $\lambda$ achieves optimal performance by balancing validation alignment with task diversity, enabling VAO to emphasize beneficial sources while preserving sufficient diversity.
Notably, minority tasks are more sensitive to $\lambda$, as their performance hinges critically on precise calibration between auxiliary data and target alignment.





\textbf{Effects of FiLM Structure.}
To evaluate the effectiveness of dynamic feature modulation in multi-task auto-bidding, we perform an ablation experiment on the FiLM module.
The results are shown in Table \ref{table_ablation_film}.
It show that the module is essential for capturing task-specific feature interactions and adapting the shared generative backbone to diverse bidding constraints.
Even without the FiLM module, our method still achieves superior performance compared to the baselines in Table \ref{table_multi_task}, which demonstrates that the validation-aligned weighting mechanism itself is a powerful driver of performance.
The inclusion of FiLM provides an additional performance boost by allowing the model to perform fine-grained conditional generation,






\section{Conclusion}
This work addresses the challenge of cross-task data sharing in generative auto-bidding.
Our theoretical analysis reveals that naive data-proportional sharing introduces a distribution shift bias that degrades minority task performance.
To mitigate this, we propose Validation-Aligned Optimization (VAO), which learns task-specific mixture weights by optimizing validation performance. 
VAO implicitly accounts for distributional alignment through gradient-based validation dynamics, ensuring that constructive knowledge is shared across tasks.
Extensive experiments demonstrate that VAO substantially outperforms SOTA baselines across multiple bidding tasks.
Our work provides both theoretical insights into cross-task data-sharing mechanisms and a practical framework for effective transfer across heterogeneous tasks.


\section*{Impact Statements}
This work proposes Validation-Aligned Optimization (VAO) to improve data efficiency and generalization in generative auto-bidding, particularly in small-scale or data-sparse advertising settings. 
By enabling principled cross-task data sharing, the approach can benefit industry practitioners by reducing data requirements, improving bidding performance for advertisers with limited participation, and lowering operational and training costs. 
As VAO operates purely at the level of optimization and training dynamics, without introducing new model capabilities or decision objectives, it does not create new societal or ethical risks beyond those inherent to online advertising systems. 
Potential downstream impacts, such as effects on market competition, ad allocation, or user experience, remain dependent on deployment context and existing platform governance.

\bibliography{icml_2026}
\bibliographystyle{icml2026}

\newpage
\onecolumn
\appendix


\section{Literature Review} \label{literature_review}
\textbf{Auto-bidding Methods.}
The mainstream auto-bidding methods can be broadly categorized into two branches to achieve diverse bidding tasks: Reinforcement Learning (RL)-based auto-bidding methods and generative auto-bidding methods.
\textit{RL-based auto-bidding methods} model auto-bidding as a Markov Decision Process and learn the optimal bidding policy through RL techniques.
For example, Deep Reinforcement Learning to Bid (DRLB) \citep{wu2018budget} uses deep Q-network (DQN) \citep{mnih2015human} with reward shaping to maximize impression value under budget constraints.
USCB \citep{He2021uscb} employs the DDPG \citep{silver2014deterministic} algorithm to dynamically adjust bidding parameters to an optimal strategy.
SORL \citep{Mou2022sorl} develops a variance-suppressed conservative Q-learning method to effectively learn auto-bidding policies.
Due to the risks of real-time bidding, offline RL methods such as BCQ \citep{fujimoto2019off}, SCAS \citep{mao2024offline}, CQL \citep{kumar2020conservative}, DMG \citep{mao2024doubly}, and IQL \citep{kostrikovoffline} have gained prominence for learning policies solely from historical datasets without online interaction.

\textit{Generative auto-bidding methods} show greater potential than RL-based methods because they can better model the complex distribution of bidding strategies. These methods adopt model-based planning strategies \citep{wang2022model} and formulate auto-bidding as a conditional generative modeling problem.
Decision Transformer (DT) \citep{Chen2021DT} generates optimal actions using an auto-regressive transformer conditioned on desired returns, historical states, and action.
GAS \citep{li2025gas} adopts DT to generate actions for auto-bidding and employs a Monte Carlo Tree Search-inspired post-training refinement to better align generated bids with diverse user preferences.
AIGB \citep{Guo2024aigb} employs a conditional diffusion model to generate bidding trajectories alongside an inverse dynamic model for action generation.
In this work, we focus on generative auto-bidding methods and build upon AIGB to address the challenges of cross-task data sharing.

\textbf{Cross-Task Data Sharing.}
Data sharing across heterogeneous tasks is a well-studied strategy for mitigating data scarcity, primarily within multi-task reinforcement learning \citep{kalashnikov2021mt,yu2022leverage,mitchell2021offline}.
These methods typically reuse experience by treating trajectories from a source task as transitions for a target task, often modifying the rewards or objectives to maintain consistency.
Direct data sharing in offline RL often fails due to significant distribution shifts. 
To address this, CDS \citep{yu2021conservative} introduces selection criteria based on conservative value differences to identify data relevant to the main task, while CDS-Zero \citep{yu2022leverage} eliminates the relabeling process by setting shared rewards to zero, thereby reducing reward bias.
UTDS \citep{bai2024pessimistic} penalizes the out-of-distribution (OOD) actions through direct OOD sampling and the associated uncertainty quantification.
While effective in standard RL, these approaches are primarily designed to address overestimation bias in scalar value functions ($Q$-learning). 
In contrast, generative auto-bidding involves matching high-dimensional trajectory distributions. 
As a result, value-based relabeling fails to capture the intricate, constraint-conditioned distributions necessary for precise bidding behavior.

\textbf{Multi-task Learning.}
Multi-task learning aims to jointly learn multiple tasks within a single model, improving learning efficiency by enabling information sharing across tasks \citep{caruana1997multitask,sun2020adashare,xu2020knowledge,thung2018brief,shen2024go4align,shen2023episodic,yang2022cross}.
Common architectural approaches include the shared bottom model, which employs a shared backbone with separate task-specific heads. 
More advanced variants include cross-stitch networks \citep{misra2016cross}, which learn adaptive feature sharing between tasks, and the Multi-Task Attention Network (MTAN) \citep{liu2019end}, which uses soft attention to dynamically select shared features for each task.
Other representative architectures are MMoE \citep{ma2018modeling} and PLE \citep{tang2020progressive}, designed to balance shared and task-specific representations effectively.
For multi-task optimization, approaches can be roughly divided into gradient-based and loss-based methods. 
\textit{Gradient-based methods} balance tasks by manipulating gradients, including Pareto optimal solutions \citep{desideri2012multiple,sener2018multi}, gradient normalization \citep{chen2018gradnorm}, gradient projection \citep{yu2020gradient,liu2021conflict,liutowards2021}, gradient sign dropout \citep{chen2020just}, and Nash bargaining solution \citep{navon2022multi}.
\textit{Loss-based methods} adaptively adjust task-specific loss weights during training to balance learning progress among tasks.
Representative approaches include uncertainty weighting \citep{kendall2018multi}, 
random loss weighting \citep{lin2021reasonable}, and strategies based on learning dynamics \citep{liu2019end,liu2023famo,shen2024go4align}.

In contrast to traditional multi-task learning (MTL) that focuses on balancing gradients or losses to avoid task interference, our work shifts the perspective toward validation-aligned data sharing. While existing gradient-based methods like PCGrad \citep{yu2020gradient} or GradNorm \citep{chen2018gradnorm} treat all training tasks as equally relevant contributors to the shared backbone, our method recognizes that in the context of generative auto-bidding, certain source tasks may introduce distribution shifts that are detrimental to the target task's performance.
By leveraging a validation signal to dynamically weight the influence of shared data, we bridge the gap between multi-task optimization and out-of-distribution robustness. 
This allows us to move beyond simple interference mitigation to a more proactive form of transfer learning, where the helpfulness of a source task is quantified by its directional alignment with the target task's generalization requirements.


\textbf{Gradient Hint and Alignment Methods.}
Our approach shares conceptual roots with a broad class of methods utilizing gradient alignment, particularly in continual learning \citep{lopez2017gradient}, online learning \citep{dekel2017online, bhaskara2020online}, and stochastic optimization \citep{baydin2017online}. 
For instance, Gradient Episodic Memory (GEM) \citep{lopez2017gradient} utilizes gradient inner products as constraints to prevent catastrophic forgetting. Similarly, in online learning, alignment signals are often treated as hints to gate updates or adapt learning rates \citep{dekel2017online, baydin2017online, bhaskara2020online}.
However, VAO departs from these works in several fundamental ways.
Unlike GEM or online hint methods that typically use alignment as a binary gate or a scalar constraint on a single update, VAO formulates data sharing as a convex combination of trajectory distributions, providing a smooth, principled way to aggregate diverse source tasks.
Furthermore, VAO introduces the validation signal to guide the transfer, ensuring that sharing is specifically aligned with the target task's generalization requirements. 
VAO is uniquely designed to mitigate the low-fidelity approximation that occurs when divergent generative distributions are naively conflated, which is a problem not addressed by traditional gradient-gating methods.
In summary, while the use of gradient alignment is an established concept, VAO’s novelty lies in its application to generative cross-task data sharing and its formulation of an entropy-weighted mechanism that explicitly navigates the distribution shifts inherent in bidding auctions.

\begin{algorithm}[t]
    \caption{Multi-Task Learning with Validation Aligned Optimization}
    \begin{algorithmic}[1]
        \STATE \textbf{Input}:  
        Maximum iteration number $T$;
        Learning rate $\eta$; 
        Temperature hyperparameter $\lambda$;
        Target task $k$;
        Validation set $\mathcal{D}_k^{\mathrm{val}}$; 
        
        \STATE Initialize model parameters ${\bm\theta}_0$; 
        \STATE Relabel source task data to target task data $\{\mathcal{D}_{i \rightarrow k}\}_{i=1}^K$;
    
        \FOR{$t = 0$ to $T$}
        \FOR{$k = 1$ to $K$} 
            \STATE Compute validation gradient $\bm g_k$ through $\mathcal{D}_k^{\mathrm{val}}$;
            \FOR{$i = 1$ to $K$} 
            
            \STATE Compute cross-task training gradient $\bm g_{ik}$ through $\mathcal{D}_{i \rightarrow k}$;
            
            \STATE Compute marginal gains $m_{ik} = \langle \bm g_k, \bm g_{ik}\rangle$;
            
            \STATE Compute weights $w_{ik} = \frac{\exp(m_{ik}/\lambda)}{\sum_{j=1}^K \exp(m_{jk}/\lambda)}$;
            \ENDFOR
            \STATE Update parameters ${\bm\theta}_{t+1} = {\bm\theta}_t - \eta \sum_i w_{ik} \bm g_{ik}$;

         \ENDFOR   
        \ENDFOR
    \end{algorithmic}
\label{algorithm_multi_task}
\end{algorithm}

\section{Pseudo Algorithm}
For a better understanding of the proposed validation alignment optimization applied in multi-task settings, we provide the Pseudo Algorithm \ref{algorithm_multi_task} in this section.
The process iteratively computes the alignment between source task training gradients and target task validation gradients, updating the distributional weights $w_{ik}$ to ensure that the generative planner prioritizes data that is directionally consistent with the target objective.

\section{Theorems \& Proofs} \label{appendix:proof}

\subsection{Assumptions}\label{app:reasonablity of assumption}
\textbf{Assumption \ref{assump:regularity}}
\textit {We assume: (i) the loss function $\ell_k(\tau; \bm{\theta})$ is bounded, i.e., $\ell_k(\tau; \bm{\theta})\in [0,M]$ for all $\tau, \bm{\theta}$; (ii) $\ell_k(\cdot; \bm{\theta})$ is $L$-Lipschitz with respect to $\bm{\theta}$; (iii) the hypothesis class $\Theta$ has finite Rademacher complexity $\mathfrak{R}_N(\Theta)$ \citep{shalev2014understanding}.}

These assumptions are standard in statistical learning and align with the practicalities of auto-bidding.

\subsection{Proof of Theorem \ref{thm:naive_bound}}
\label{proof:naive_generalization_bound}
{\textbf{Theorem \ref{thm:naive_bound}} (Generalization Bound under Naive Cross-Task Data Sharing).
\textit{Under Assumption \ref{assump:regularity}, with probability at least $1-\delta$, we have the following generalization bound in the presence of naive cross-task data sharing:
\begin{equation}
\begin{aligned}
    \mathcal{R}_k(\bm{\theta}) \leq \hat{\mathcal{R}}_{k}^\mathrm{Naive
}(\bm{\theta}) 
+  \underbrace{M \sum_{i \neq k} \alpha_i d_{\mathrm{TV}}(\mathbb{P}_i, \mathbb{P}_k)}_{\text{distribution shift bias}}
+  \underbrace{2L \, \mathfrak{R}_{N}(\Theta) 
+ M \sqrt{\frac{\log(2/\delta)}{2N}}}_{\text{estimation error}},
\end{aligned}
\end{equation}
where $d_{\mathrm{TV}}(\mathbb{P}_i, \mathbb{P}_k)=\sup_A|\mathbb{P}_i(A)-\mathbb{P}_k(A)|$ denotes the total variation distance between task distributions. 
}

\begin{proof}
    $\mathcal{R}_k(\bm{\theta}) - \hat{\mathcal{R}}_{k}^\mathrm{Naive
}(\bm{\theta})$ can be decomposed to two parts, i.e., 
\begin{equation}
    \mathcal{R}_k(\bm{\theta}) - \hat{\mathcal{R}}_{k}^\mathrm{Naive
}(\bm{\theta})=\underbrace{\mathcal{R}_k(\bm{\theta})- {\mathcal{R}}^{\mathrm{Naive}}_k(\bm{\theta})}_{(\mathrm{I})}+\underbrace{  {\mathcal{R}}^{\mathrm{Naive}}_k(\bm{\theta}) - \hat{\mathcal{R}}_{k}^{\mathrm{Naive}}(\bm{\theta})}_{(\mathrm{II})}.
\end{equation}

The first term ($\mathrm{I}$) represents the error between the true target risk and the naive mixed empirical risk.
Recall the definition of ${\mathcal{R}}_k^{\mathrm{Naive}}(\bm{\theta}) = \sum \alpha_i {\mathcal{R}}_i(\bm{\theta})$, where $\mathcal{R}_i(\bm{\theta})=\mathbb{E}_{(\tau,\bm y_\tau)\sim\mathbb{P}_i}[\ell_k(\tau;\bm \theta)]$  evaluates the target loss $\ell_k$ on trajectories from source distribution $\mathbb{P}_i$.
Using the fact that $\sum_{i=1}^K \alpha_i=1$, we have:
\begin{equation}
\begin{aligned}
    \mathcal{R}_k(\bm{\theta})- {\mathcal{R}}^{\mathrm{Naive}}_k(\bm{\theta})
    &= \mathcal{R}_k(\bm{\theta})-\sum_{i=1}^K \alpha_i {\mathcal{R}}_i(\bm{\theta})
    = \sum_{i=1}^K \alpha_i \mathcal{R}_k(\bm{\theta}) - \sum_{i=1}^K \alpha_i {\mathcal{R}}_i(\bm{\theta})\\
    &= \sum_{i=1}^K \alpha_i \left(\mathcal{R}_k(\bm{\theta}) - \mathcal{R}_i(\bm{\theta})\right)
    = \sum_{i\neq k}\alpha_i\left(\mathcal{R}_k(\bm{\theta})-\mathcal{R}_i(\bm{\theta})\right).
\end{aligned}
\end{equation}

Expanding the expectations:
\begin{equation}
\begin{aligned}
    \mathcal{R}_k(\bm{\theta})- {\mathcal{R}}^{\mathrm{Naive}}_k(\bm{\theta})=\sum_{i\neq k} \alpha_i \left( \mathbb{E}_{(\tau,\bm y_\tau)\sim\mathbb{P}_k}[\ell_k(\tau;\bm \theta)] - \mathbb{E}_{(\tau,\bm y_\tau)\sim\mathbb{P}_i}[\ell_k(\tau;\bm \theta)] \right).
\end{aligned}
\end{equation}
For any bounded loss $\ell_k\in [0,M]$ and any two probability distributions $\mathbb{P}$, $\mathbb{Q}$, it holds that \citep{sriperumbudur2009integral}:
\begin{equation}
    \bigl| \mathbb{E}_{\mathbb{P}}[\ell] - \mathbb{E}_{\mathbb{Q}}[\ell] \bigr| \leq M \cdot d_{\mathrm{TV}}(\mathbb{P}, \mathbb{Q}),
\end{equation}
where $d_{\mathrm{TV}}(\mathbb{P}, \mathbb{Q})=\sup_{A}|\mathbb{P}(A)-\mathbb{Q}(A)|=\frac{1}{2}\|\mathbb{P}-\mathbb{Q}\|_1$. 
Applying this bound to each term $i\neq k$ yields
\begin{equation}
    \mathcal{R}_k(\bm{\theta})- {\mathcal{R}}^{\mathrm{Naive}}_k(\bm{\theta})\leq M\sum_{i\neq k} \alpha_i d_{\mathrm{TV}}(\mathbb{P}_i, \mathbb{P}_k).
\end{equation}
The second term ($\mathrm{II}$) captures the standard generalization gap between the population naive mixture and its empirical counterpart.
Under the assumption that the loss $\ell_k$ is bounded in $[0,M]$ and $L$-Lipschitz with respect to $\bm \theta$, we apply typical results based on Rademacher complexity \citep{mohri2018foundations}.
With probability at least $1-\delta$,
\begin{equation}
    \sup_{\bm{\theta} \in \Theta} \big| \mathcal{R}^{\mathrm{Naive}}_k(\bm{\theta}) - \hat{\mathcal{R}}^{\mathrm{Naive}}_k(\bm{\theta}) \big|
\leq 2L \mathfrak{R}_{N}(\Theta) + M \sqrt{\frac{\log(2/\delta)}{2N}},
\end{equation}
where $N=\sum_{i=1}^KN_i$ is the total number of samples across all tasks, and $\mathfrak{R}_{N}(\Theta)$ denotes the Rademacher complexity of the hypothesis class $\Theta$ with respect to the pooled dataset of size $N$.
The factor $2L$ arises from the Lipschitz constant of the loss, and the concentration term $M\sqrt{\log(2/\delta)/(2N)}$ follows from Hoeffding's inequality applied to the bounded loss function.

Combining the bounds from the above steps, we obtain
\begin{equation}
\mathcal{R}_k(\bm{\theta}) \leq \hat{\mathcal{R}}_{k}^\mathrm{Naive
}(\bm{\theta}) 
+ M \sum_{i \neq k} \myheight \alpha_i \myheight d_{\mathrm{TV}}(\mathbb{P}_i, \mathbb{P}_k)
+ 2L \, \mathfrak{R}_{N}(\Theta) 
+ M \sqrt{\frac{\log(2/\delta)}{2 N}}.
\end{equation}
\end{proof}

\subsection{Proof of Theorem \ref{thm:improvement_guaranteen}}
\label{proof:thm:improvement_guaranteen}
{\textbf{Theorem \ref{thm:improvement_guaranteen}} (Improvement Guarantee of VAO).
\textit{Under the same setting, let $\bm{\theta}_{t+1}^{\mathrm{VAO}}$ and $\bm{\theta}_{t+1}^{\mathrm{Naive}}$ denote the parameters updated after one gradient step from $\bm \theta_t$ with VAO weights $\{w_{ik}^*\}$ and naive weights $\{\alpha_i\}$, respectively. Then, we have
\begin{equation}
\begin{aligned}
    \mathcal{R}_k(\bm{\theta}_{t+1}^{\mathrm{VAO}})  \leq \mathcal{R}_k(\bm{\theta}_{t+1}^{\mathrm{Naive}})
     - \eta \Delta + \mathcal{O}(\eta^2), \quad\quad
     \Delta =\sum_{i=1}^K (w_{ik}^* - \alpha_i) m_{ik},
\end{aligned}
\end{equation}
where $\Delta \geq 0$, and $\Delta > 0$ whenever the naive weights $\{\alpha_i\}$ deviate from the alignment-optimized weights $\{w_{ik}^*\}$.
}

\begin{proof}
We analyze the improvement by comparing the validation risk after one gradient step under VAO versus naive sharing.
Starting from parameters $\bm \theta_t$ at iteration $t$, the gradient update under naive sharing produces:
\begin{equation}
    \bm{\theta}_{t+1}^{\mathrm{Naive}} = \bm{\theta}_t - \eta \sum_{i=1}^K \alpha_i \bm{g}_{ik}.
\end{equation}
Similarly, under VAO, the update is 
\begin{equation}
    \bm{\theta}_{t+1}^{\mathrm{VAO}} = \bm{\theta}_t - \eta \sum_{i=1}^K w_{ik}^* \bm{g}_{ik}.
\end{equation}
The VAO weights $\{w_{ik}^*\}$ are obtained by minimizing the validation risk change in Eq. \eqref{eq:regularized_obj}:
\begin{equation}
    \{w_{ik}^*\} = \arg\max_{\{w_{ik}\}: \sum_i w_{ik}=1, w_{ik}\geq 0} \sum_{i=1}^K w_{ik} m_{ik}-\lambda \sum_{i=1}^Kw_{ik}\ln w_{ik},
\end{equation}

Using a first-order Taylor expansion of the risk $\mathcal{R}_k$ around $\bm{\theta}_t$, we approximate the risk after one update step as:
\begin{equation}
\mathcal{R}_k(\bm{\theta}_{t+1}) = \mathcal{R}_k(\bm{\theta}_t) + \langle \nabla_{\bm \theta}\mathcal{R}_k(\bm{\theta}_t), \bm{\theta}_{t+1} - \bm{\theta}_t \rangle + \mathcal{O}(\eta^2).
\end{equation}
Substituting the update rules for $\bm{\theta}_{t+1}^{\mathrm{Naive}}$ and $\bm{\theta}_{t+1}^{\mathrm{VAO}}$, and noting that the validation gradient $\bm{g}_k = \nabla_{\bm \theta} \mathcal{R}_k(\bm{\theta}_t)$, we have:
\begin{align}
\mathcal{R}_k(\bm{\theta}_{t+1}^{\mathrm{Naive}}) &= \mathcal{R}_k(\bm{\theta}_t) - \eta \sum_{i=1}^K \alpha_i \langle \bm{g}_k, \bm{g}_{ik} \rangle + \mathcal{O}(\eta^2) = \mathcal{R}_k(\bm{\theta}_t) - \eta \sum_{i=1}^K \alpha_i m_{ik} + \mathcal{O}(\eta^2), \\
\mathcal{R}_k(\bm{\theta}_{t+1}^{\mathrm{VAO}}) &= \mathcal{R}_k(\bm{\theta}_t) - \eta \sum_{i=1}^K w_{ik}^* m_{ik} + \mathcal{O}(\eta^2).
\end{align}

Subtracting the naive risk from the VAO risk yields: 
\begin{equation} 
\mathcal{R}_k(\bm{\theta}_{t+1}^{\mathrm{VAO}}) - \mathcal{R}_k(\bm{\theta}_{t+1}^{\mathrm{Naive}}) = - \eta \sum_{i=1}^K (w_{ik}^* - \alpha_i) m_{ik} + \mathcal{O}(\eta^2). 
\end{equation}
Let $\Delta = \sum_{i=1}^K (w_{ik}^* - \alpha_i) m_{ik}$.
To show $\Delta \geq 0$, let $f(\bm{w}) = \sum_i w_{ik} m_{ik} - \lambda \sum_i w_{ik} \ln w_{ik}$ be the VAO objective function. 
Since $\bm{w}^*$ is the maximizer of $f(\bm{w})$ over the simplex, and the naive weights $\bm{\alpha}$ also lie in the simplex (where $\sum \alpha_i = 1, \alpha_i \geq 0$), it follows by the definition of a maximizer that:
\begin{equation}
f(\bm{w}^*) \geq f(\bm{\alpha}),
\end{equation}
i.e., 
\begin{equation}
    \sum_{i=1}^K w_{ik}^* m_{ik} - \lambda \sum_{i=1}^K w_{ik}^* \ln w_{ik}^* \geq \sum_{i=1}^K \alpha_i m_{ik} - \lambda \sum_{i=1}^K \alpha_i \ln \alpha_i.
\end{equation}
Rearrange
\begin{equation}
    \Delta = \sum_{i=1}^K (w_{ik}^* - \alpha_i) m_{ik} \geq \lambda \sum_{i=1}^K (w_{ik}^* \ln w_{ik}^*-\alpha_i \ln \alpha_i).
\end{equation}
Consider the case where $\lambda=0$, the inequality $\Delta\geq0$ always holds.
For $\lambda>0$, the regularization
smooths the weight distribution, but the fundamental property remains: the optimization ensures that $\bm{w}^*$ achieves at least as high a weighted alignment as any other feasible weights, including $\bm{\alpha}$.
The bound $\Delta \geq 0$ holds because the optimization of $f(\bm{w})$ jointly considers both alignment maximization and entropy regularization.
\end{proof}


\section{Implementation Details}
The statistics of the dataset are summarized in Table \ref{tab:data_split}.
The generative planner is implemented as a Causal Transformer to capture the dependencies within bidding trajectories.
We use a batch size of 32 and a learning rate of $1e-4$ with the Adam optimizer. 
The state dimension is 17, encompassing auction features, historical winning prices, and real-time budget status, while the action dimension is 1, representing the bidding scalar.
To ensure transition feasibility, an inverse dynamic model is trained as a 3-layer MLP. This model maps $(\bm s_t, \bm s_{t+1})$ to the predicted action $a_t$, serving as a consistency regularizer.
For gradient alignment, we utilize cosine similarity to compute $m_{ik}$, removing the influence of gradient magnitude and focusing on directional consistency.  
To maintain computational efficiency, VAO weights are updated every 5 training steps. 
This is a reasonable optimization because the relative alignment between task distributions evolves slowly relative to the parameter updates. 
This sparse update schedule ensures that the computational overhead of calculating validation gradients remains a negligible fraction of the total training time, maintaining high efficiency without sacrificing the learned weight quality.
The experiments are conducted based on an NVIDIA T4 Tensor Core GPU. We use 10 CPUs and 200G memory. 

\begin{table}[t]
    \centering
    \caption{\textbf{Statistics of datasets.}}
    \small
    \begin{tabular}{c|cccc}
    \toprule
        \multirow{2}{*}{Task}  & \multicolumn{2}{c}{Train Traj.}  & \multirow{2}{*}{Val Traj.} & \multirow{2}{*}{Test Traj.}  \\
        \cmidrule(lr){2-3}
        & No Sharing & Sharing && \\
        \midrule
         CPA-MCB&  4,160 & +95,680 & 20 & 20 \\
         TargetROAS&  12,480 & +87,360 & 60 & 60 \\
         BCB& 83,200 & +16,640 & 400 & 400 \\
         \bottomrule
    \end{tabular}
    \label{tab:data_split}
\end{table}

\section{Additional Experimental Results}\label{app:exp}


    

In addition to the AIGB implementation in the main paper, we also evaluate our method on AIGB-Pearl \citep{mou2025enhancing}, a state-of-the-art generative auto-bidding approach. 
Results on both single-task and multi-task settings are shown in Tables \ref{append_table_single_task} and \ref{append_table_multi_task}, demonstrating VAO yields consistent gains across all tasks.
These results demonstrate that VAO is agnostic to the underlying generative backbone, successfully navigating distribution shifts whether applied to standard causal architectures or more complex models.
By focusing on gradient-level alignment rather than specific architectural constraints, VAO serves as a plug-and-play enhancement that improves data-scarce task performance regardless of the specific auto-bidding architecture employed. This versatility suggests that VAO can be seamlessly integrated into a wide array of existing generative pipelines to solve the inherent challenges of cross-task data sharing.





\setlength{\tabcolsep}{9.0pt}
\begin{table*}[t]
\caption{\textbf{Performance comparison with baselines on three bidding tasks under single-task learning settings.} Mean and standard error are reported across five seeds. Metrics include value and score for each task. 
\textbf{Bold} denotes the best results.
$\uparrow$ denotes the higher the better.}
\small
\centering
\begin{tabular}{ll|cccccc}
\toprule
 \multicolumn{2}{c|}{\multirow{2}{*}{\textbf{Method}}} & \multicolumn{2}{c}{\textbf{CPA-MCB}}  & \multicolumn{2}{c}{\textbf{TargetROAS}}  & \multicolumn{2}{c}{\textbf{BCB}} \\
\cmidrule(lr){3-4}\cmidrule(lr){5-6}\cmidrule(lr){7-8}
\multicolumn{2}{c|}{~} & Value $\uparrow$ & Score $\uparrow$ & Value $\uparrow$ & Score $\uparrow$ & Value $\uparrow$ & Score $\uparrow$\\
\midrule
MDMM & No Sharing & 30.38\scriptsize{$\pm$0.20} & 23.02\scriptsize{$\pm$0.25} & 26.98\scriptsize{$\pm$0.09}& 27.43\scriptsize{$\pm$0.07} & 25.19\scriptsize{$\pm$0.12} & 20.61\scriptsize{$\pm$0.16} \\
\midrule
\multirow{2}{*}{ODMM} & Naive Sharing & 28.27\scriptsize{$\pm$0.31} & 20.99\scriptsize{$\pm$0.69} &29.13\scriptsize{$\pm$0.27}  & 29.91\scriptsize{$\pm$0.31} &24.39\scriptsize{$\pm$0.08} & 19.94\scriptsize{$\pm$0.08}  \\
&  \cellcolor{myhighlight}   \textbf{VAO (Ours)} &\cellcolor{myhighlight}\textbf{33.46\scriptsize{$\pm$0.44}} &
\cellcolor{myhighlight}\textbf{25.60\scriptsize{$\pm$0.72}} & \cellcolor{myhighlight}\textbf{30.01\scriptsize{$\pm$0.20}} & \cellcolor{myhighlight}\textbf{30.96\scriptsize{$\pm$0.22}} & \cellcolor{myhighlight}\textbf{27.46\scriptsize{$\pm$0.25}} & \cellcolor{myhighlight}\textbf{23.87\scriptsize{$\pm$0.26}}
\\
\bottomrule
\end{tabular}
\label{append_table_single_task}
\end{table*}

\setlength{\tabcolsep}{4.8pt}
\begin{table*}[t]
\caption{\textbf{Performance comparison with baselines on three bidding tasks under multi-task learning settings.} Mean and standard error are reported across five seeds. Metrics include value and score for each task and overall MTL performance $\Delta m\%$. \textbf{Bold} and \underline{underlined} denote the best and the most competitive results. $\downarrow$ denotes the lower the better.}
\small
\centering
\begin{tabular}{ll|cccccc|c}
\toprule
\multicolumn{2}{c|}{\multirow{2}{*}{\textbf{Method}}} & \multicolumn{2}{c}{\textbf{CPA-MCB}}  & \multicolumn{2}{c}{\textbf{TargetROAS}}  & \multicolumn{2}{c|}{\textbf{BCB}} & \multirow{2}{*}{$\Delta m\% \downarrow$}\\
\cmidrule(lr){3-4}\cmidrule(lr){5-6}\cmidrule(lr){7-8}
\multicolumn{2}{c|}{~} & Value $\uparrow$ & Score $\uparrow$ & Value $\uparrow$ & Score $\uparrow$ & Value $\uparrow$ & Score $\uparrow$\\
\midrule
MDMM & STL (No Sharing) &30.38\scriptsize{$\pm$0.20} & 23.02\scriptsize{$\pm$0.25} & 26.98\scriptsize{$\pm$0.09}& 27.43\scriptsize{$\pm$0.07} & 25.19\scriptsize{$\pm$0.12} & 20.61\scriptsize{$\pm$0.16}& - \\
\midrule
\multirow{4}{*}{MDOM} & EW & 31.39\scriptsize{$\pm$0.40} & 25.30\scriptsize{$\pm$0.50} & 27.34\scriptsize{$\pm$0.23} & 27.87\scriptsize{$\pm$0.25}& 25.10\scriptsize{$\pm$0.20} & 20.72\scriptsize{$\pm$0.24} & -2.72 \\
& FAMO \citep{liu2023famo} & 33.48\scriptsize{$\pm$0.97} & 26.50\scriptsize{$\pm$0.82}& 28.98\scriptsize{$\pm$0.58} & 29.61\scriptsize{$\pm$0.65}& 26.52\scriptsize{$\pm$0.57} & 22.46\scriptsize{$\pm$0.69}& -9.15 \\
& PCGrad \citep{yu2020gradient} & \underline{38.90\scriptsize{$\pm$0.07}} & \underline{32.04\scriptsize{$\pm$0.17}} & 32.77\scriptsize{$\pm$0.04} & \underline{33.52\scriptsize{$\pm$0.03}}& 29.53\scriptsize{$\pm$0.03} &26.22\scriptsize{$\pm$0.04} & -25.89 \\
& FairGrad \citep{ban2024fair} & 27.95\scriptsize{$\pm$0.42} & 25.15\scriptsize{$\pm$0.39}& 26.54\scriptsize{$\pm$0.35} & 25.49\scriptsize{$\pm$0.46}& 22.70\scriptsize{$\pm$0.18} & 17.94\scriptsize{$\pm$0.17}& 5.05 \\
\midrule
\multirow{5}{*}{ODOM} & EW & 33.29\scriptsize{$\pm$0.32} & 27.71\scriptsize{$\pm$0.40}& 29.25\scriptsize{$\pm$0.16} & 29.94\scriptsize{$\pm$0.18}& 27.73\scriptsize{$\pm$0.37} & 22.30\scriptsize{$\pm$0.38}& -10.97 \\
& FAMO \citep{liu2023famo} & 35.46\scriptsize{$\pm$0.23} & 28.35\scriptsize{$\pm$0.22}& 30.36\scriptsize{$\pm$0.11} & 31.21\scriptsize{$\pm$0.11}& 27.56\scriptsize{$\pm$0.12}& 23.77\scriptsize{$\pm$0.16} & -15.15 \\
& PCGrad \citep{yu2020gradient} & 38.86\scriptsize{$\pm$0.38} & 31.57\scriptsize{$\pm$0.28}& \underline{32.85\scriptsize{$\pm$0.22}}  & 33.48\scriptsize{$\pm$0.16}& \underline{29.81\scriptsize{$\pm$0.17}} & \underline{26.53\scriptsize{$\pm$0.27}}& \underline{-25.99} \\
& FairGrad \citep{ban2024fair} & 31.74\scriptsize{$\pm$0.29} & 25.39\scriptsize{$\pm$0.25}& 29.90\scriptsize{$\pm$0.26} &30.03\scriptsize{$\pm$0.12} & 28.12\scriptsize{$\pm$0.25} & 23.94\scriptsize{$\pm$0.15} & -10.48  \\
 & \cellcolor{myhighlight}\textbf{VAO (Ours)} &\cellcolor{myhighlight}\textbf{39.15\scriptsize{$\pm$0.25}} &
\cellcolor{myhighlight}\textbf{32.67\scriptsize{$\pm$0.25}} & \cellcolor{myhighlight}\textbf{33.42\scriptsize{$\pm$0.25}} & \cellcolor{myhighlight}\textbf{33.88\scriptsize{$\pm$0.25}} & \cellcolor{myhighlight}\textbf{30.12\scriptsize{$\pm$0.25}} & \cellcolor{myhighlight}\textbf{27.04\scriptsize{$\pm$0.25}}& \cellcolor{myhighlight}\textbf{-28.16}\\
\bottomrule
\end{tabular}
\label{append_table_multi_task}
\vspace{-2mm}
\end{table*}

\section{Complexity Analysis}
To evaluate the computational overhead of VAO, we analyze its complexity relative to naive cross-task sharing. 
The primary additional cost stems from the computation of the alignment weights $w_{ik}$, which involves calculating the inner product $m_{ik}$ between training and validation gradients.
For a model with $D$ parameters and $K$ source tasks, this introduces a computational complexity of $\mathcal{O}(KD)$ per weight update.
While this might appear significant, our approach remains highly efficient in practice. 
In our implementation, we update the task weights $w_{ik}^*$ every $T$ training steps (e.g., $T=5$) rather than at every iteration. 
This optimization is justified because gradient alignments evolve on a much slower manifold than the model parameters themselves. 
Furthermore, decoupling the weight updates from every iteration prevents the introduction of unnecessary variance from stochastic gradient estimation, ensuring more stable weight quality. 
As a result, VAO achieves superior performance with only a marginal increase in total training time, maintaining the scalability required for large-scale advertising environments.



\end{document}